\documentclass{article}

\PassOptionsToPackage{numbers, compress}{natbib}
\usepackage[main,final]{neurips_2026}

\usepackage{algorithm}
\usepackage{algorithmic}  % 或者 algorithmicx / algpseudocode
\usepackage{graphicx}
\usepackage[utf8]{inputenc} % allow utf-8 input
\usepackage[T1]{fontenc}    % use 8-bit T1 fonts
\usepackage{hyperref}       % hyperlinks
\usepackage{url}            % simple URL typesetting
\usepackage{booktabs}       % professional-quality tables
\usepackage{amsmath}
\usepackage{amssymb}
\usepackage{mathtools}
\usepackage{amsthm}
\usepackage{nicefrac}       % compact symbols for 1/2, etc.
\usepackage{microtype}      % microtypography
\usepackage{xcolor}         % colors
\usepackage{subcaption}
\usepackage{wrapfig}
\usepackage{placeins}
\makeatletter
\theoremstyle{plain}
\newtheorem{theorem}{Theorem}[section]
\newtheorem{proposition}[theorem]{Proposition}
\newtheorem{lemma}[theorem]{Lemma}

\theoremstyle{definition}

\theoremstyle{remark}

\newcommand{\bx}{\mathbf{x}}
\newcommand{\bz}{\mathbf{z}}
\newcommand{\f}{\mathbf{f}}
\newcommand{\vvec}{\mathbf{v}}    % Renamed from \v to avoid conflict with accents
\newcommand{\uvec}{\mathbf{u}}    % Renamed from \u to avoid conflict with accents
\newcounter{myalg}
\makeatother

\title{Flow Perturbation++: Multi-Step Unbiased Jacobian Estimation for High-Dimensional Boltzmann Sampling}

\author{%
  Xin Peng$^{1,2}$ \quad
  Ang Gao$^{1}$\thanks{Corresponding author} \\
  $^{1}$State Key Laboratory of Information Photonics and Optical Communications, \\
  School of Physical Science and Technology, \\
  Beijing University of Posts and Telecommunications (BUPT) \\
  $^{2}$Shanghai Academy of AI for Science (SAIS) \\
  \texttt{xinpeng@bupt.edu.cn} \quad \texttt{anggao@bupt.edu.cn}
}

\begin{document}

\maketitle

\begin{abstract}
The scalability of continuous normalizing flows (CNFs) for unbiased Boltzmann sampling remains limited in high-dimensional systems due to the cost of Jacobian-determinant evaluation, which requires $D$ backpropagation passes through the flow layers. Existing stochastic Jacobian estimators such as the Hutchinson trace estimator reduce computation but introduce bias, while the recently proposed Flow Perturbation method is unbiased yet suffers from high variance. We present \textbf{Flow Perturbation++}, a variance-reduced extension of Flow Perturbation that discretizes the probability-flow ODE and performs unbiased stepwise Jacobian estimation at each integration step. This multi-step construction retains the unbiasedness of Flow Perturbation while achieves substantially lower estimator variance. Integrated into a Sequential Monte Carlo framework, Flow Perturbation++ achieves significantly improved equilibrium sampling on a 1000D Gaussian Mixture Model and the all-atom Chignolin protein compared with Hutchinson-based and single-step Flow Perturbation baselines.
\end{abstract}

\section{Introduction}
\label{sec:introduction}

Sampling from the Boltzmann distribution
\(
p(\mathbf{x}) \propto e^{-u(\mathbf{x})}
\)
is a long-standing challenge in statistical mechanics and molecular modeling.
Rugged, high-dimensional energy landscapes with numerous metastable states are separated by large free-energy barriers~\cite{frauenfelder1991energy}, causing classical Molecular Dynamics (MD)~\cite{verlet1967computer} and Markov Chain Monte Carlo (MCMC)~\cite{metropolis1953equation} methods to mix extremely slowly.
Enhanced sampling techniques---such as replica exchange~\cite{hukushima1996exchange,swendsen1986replica}, umbrella sampling~\cite{torrie1977nonphysical}, metadynamics~\cite{laio2002escaping}, and transition-path sampling~\cite{dellago1998transition}---can mitigate these difficulties but often introduce substantial computational overhead or require carefully engineered, system-specific collective variables.

Beyond physical systems, Boltzmann distributions appear as a common functional form in statistical machine learning and reinforcement learning (RL)~\cite{messaoud2024s,pmlr-v80-haarnoja18b,chao2024maximum}. In RL, such forms arise in entropy-regularized and inference-based formulations, e.g., maximum-entropy RL induces policies \(\pi(a \mid s) \propto \exp\!\big(Q^\pi(s,a)/\alpha\big)\)~\cite{haarnoja2017reinforcement, shi2019soft,jain2024sampling}, and trajectory-level methods define \(\pi(\mathbf{x}) \propto \exp\!\big(\beta\,R(\mathbf{x})\big)\)~\cite{bengio2023gflownet,malkin2022trajectory,zhu2025flowrl}. However, in these settings this form is mainly descriptive rather than used for explicit Boltzmann sampling. The prevalence of Boltzmann distributions motivates the development of efficient methods for sampling from these distributions.

This widespread demand has driven the development of deep generative models for learning Boltzmann target distributions and enabling efficient sampling.
By transporting a simple base distribution to the target, such models have achieved notable success in molecular and material design, including drug-like molecule generation, protein conformations, and novel materials~\cite{Zheng:2024ww,Abramson:2024vm,xu2022geodiff,pmlr-v162-hoogeboom22a,wang2024generating,noe2019boltzmann,tan2026amortized}.

Normalizing flows (NFs)~\cite{10.5555/3045118.3045281,dinh2015nice,dinh2017density,kingma2018glow,tan2026amortized} provide exact likelihood evaluation through invertible transformations between a simple prior and a complex target distribution.
Discrete NFs rely on carefully designed architectures to make Jacobian determinants tractable, trading expressive flexibility for computational efficiency.

Continuous normalizing flows (CNFs)~\cite{kohler2020equivariant}, a continuous-time limit of NFs, parameterize transformations via neural ordinary differential equations (neural ODEs)~\cite{chen2018neural} and have recently been applied to Boltzmann sampling~\cite{noe2019boltzmann,klein2023equivariant, Wang:2024vq,dern2026energyweightedflowmatchingunlocking}.
A key advantage of CNFs is their ease of training, enabled by simulation-free objectives such as diffusion-based probability-flow ODEs~\cite{ho2020denoising,song2020score} and flow matching~\cite{lipman2023flow, liu2023rectified, albergo2023stochastic}, which have achieved strong empirical performance in image generation and molecular modeling.

Despite these advantages, unbiased Boltzmann sampling with CNFs remains challenging. 
To obtain unbiased expectations from the Boltzmann target, one must correct the flow-induced proposal distribution via importance reweighting,
\(w(\mathbf{x}) = p(\mathbf{x}) / q(\mathbf{x})\),
which requires evaluating likelihood ratios.
Evaluating importance weights requires integrating the trace of the Jacobian of the velocity field along the flow trajectory, necessitating \(D\) backpropagation passes through all layers of the flow, where \(D\) is the dimensionality of the system~\cite{noe2019boltzmann}.
While stochastic trace estimators such as the Hutchinson estimator~\cite{hutchinson1989stochastic} reduce this cost~\cite{Wang:2024vq}, exponentiating their stochastic trace estimates can introduce convergence issues and systematic bias~\cite{noe2019boltzmann,erives2024verletflowsexactlikelihoodintegrators}.
Consequently, existing CNF-based methods have only achieved unbiased Boltzmann sampling for relatively low-dimensional systems~\cite{klein2023equivariant}.

Building on Flow Perturbation (FP)~\cite{peng2025flow}, which estimates flow Jacobians via stochastic perturbations to reduce the cost of unbiased Boltzmann sampling, we introduce \textbf{Flow Perturbation++ (FP++)}, a theoretically grounded, determinant-unbiased estimator for continuous-flow dynamics under exact inverse-step derivatives, with variance reduction under the conditions in Appendix~\ref{sec:fpplusplus_variance}. FP++ exploits the semigroup property of flow maps to decompose the global Jacobian
\(
|\det J_{1\to T}| = \prod_{k=1}^{T} |\det J_k|,
\)
estimating each factor without computing the full Jacobian directly.

Our main contributions are:
\begin{itemize}
\item \textbf{Unbiased determinant estimation.} FP++ retains the determinant-level unbiasedness of FP~\cite{peng2025flow} for exact inverse-step maps and directional derivatives, while improving numerical stability.

\item \textbf{Variance reduction.} Under the conditions in Appendix~\ref{sec:fpplusplus_variance}, decomposing the global Jacobian into per-step determinants reduces log-domain variance by removing probe-induced correlations between steps, especially for flows with many integration steps.

\item \textbf{Seamless SMC integration.} FP++ naturally integrates with Sequential Monte Carlo (SMC) annealing~\cite{liu1998sequential}, supporting stable Boltzmann reweighting and mitigating particle degeneracy.

\item \textbf{Scalable validation.} Experiments on a 1000-dimensional Gaussian mixture and an all-atom Chignolin mutant show FP++ improves equilibrium sampling over Hutchinson and original FP baselines.
\end{itemize}

\section{Related work}
\label{sec:related_work}

\subsection{Balancing expressivity and tractability in flows}
Generative models for Boltzmann sampling differ primarily in how they handle the Jacobian determinant.

\paragraph{Constrained architectures.}
Early flow models such as NICE~\cite{dinh2015nice}, RealNVP~\cite{dinh2017density}, and Glow~\cite{kingma2018glow} use structured transformations (e.g., affine coupling) that yield triangular Jacobians and allow $\mathcal{O}(D)$ determinant evaluation. While computationally efficient, these architectural constraints limit expressivity and are often insufficient for complex molecular or high-dimensional target distributions~\cite{noe2019boltzmann}.

\paragraph{Free-form continuous flows.}
More flexible approaches—including Neural ODEs~\cite{chen2018neural}, diffusion-based Probability Flows~\cite{ho2020denoising,song2020score}, flow-matching methods~\cite{lipman2023flow, liu2023rectified, albergo2023stochastic}, and i-ResNets~\cite{pmlr-v97-behrmann19a}—remove structural
constraints to model richer dynamics. i-ResNets enforce invertibility via residual blocks and estimate determinant product with Hutchinson’s trace estimator, while diffusion and flow-matching flows learn vector fields without explicit Jacobian computations during training. However, exact likelihood and unbiased inference still require evaluating Jacobian determinants, leaving a fundamental trade-off between expressivity and tractability.

\paragraph{Complementary sampler pipelines.}
FALCON learns few-step invertible maps~\cite{rehman2025falcon}, whereas FAB and SBG-SMC improve learned proposals, annealing, or sequential weighting~\cite{midgley2023fab,tan2025sbg}. These methods address proposal construction and particle transport, while their likelihood or importance weights still depend on a Jacobian correction. They therefore form a parallel and complementary line to Jacobian and likelihood estimators such as Hutchinson and FP~\cite{hutchinson1989stochastic,peng2025flow}. FP++ belongs to the latter line, which is the focus of our direct comparisons.

\subsection{Sequential Monte Carlo}
High-dimensional importance sampling often suffers from weight collapse. Sequential Monte Carlo (SMC)~\cite{liu1998sequential} mitigates this via tempering and resampling.  
Flow-SMC approaches~\cite{arbel2021annealed,speich2021sequential} combine learned flows with SMC to stabilize weight evolution.  
FP++ integrates naturally with SMC annealing, providing a determinant-based Boltzmann reweighting correction in high dimensions.
\section{Background} 

\subsection{Boltzmann reweighting}
Given a generative model $q(\bz)$ induced by a mapping $\f: \bz \mapsto \bx$ ($\bz \sim q(\bz)$), the unbiased Boltzmann expectation of an observable $O(\bx)$ is
\begin{equation}
    \mathbb{E}_{\bx \sim p}[O] =
    \frac{\mathbb{E}_{\bz \sim q(\bz)}\big[O(\f(\bz))\,e^{-W(\bz)}\big]}
         {\mathbb{E}_{\bz \sim q(\bz)}\big[e^{-W(\bz)}\big]},
\end{equation}
where the generalized work is $W(\bz) = u_X(\f(\bz)) - u_Z(\bz) - \Delta S(\bz)$, with $u_X(\bx)$ and $u_Z(\bz)$ denoting the dimensionless potentials of the target Boltzmann and prior distributions, respectively. 
The entropic contribution $\Delta S(\bz) = \log \left| \det\!\left( \frac{\partial  \f}{\partial \bz} \right) \right|$ is determined by the Jacobian of the generative map and represents the primary computational bottleneck. In practice, the importance weights $e^{-W(\bz)}$ are normalized.

\subsection{Probability flow ODEs}
Diffusion-based generative models define an SDE $d\mathbf{x}_t = b(\mathbf{x}_t,t)\,dt + g(t)\,d\mathbf{w}_t$. The corresponding probability flow ODE~\cite{song2020score} is
\begin{equation}
    \frac{d\mathbf{x}_t}{dt} = b(\mathbf{x}_t,t) - \frac{1}{2} g(t)^2 \nabla_{\mathbf{x}} \log p_t(\mathbf{x}_t),
\end{equation}
which defines a special type of continuous normalizing flow (CNF). During training, score-matching allows the model to avoid direct computation of the Jacobian determinant, which would be prohibitively expensive in high dimensions.

At inference time, evaluating importance weights or likelihoods requires the log-determinant of the flow Jacobian~\cite{kohler2020equivariant}:
\begin{equation}
    \log \left| \det \left( \frac{\partial \f}{\partial \bz} \right) \right|
    = \int_0^T \nabla_{\mathbf{x}} \cdot v(\mathbf{x}_t, t)\, dt,
\end{equation}
where $v(\mathbf{x}_t, t) = b(\mathbf{x}_t,t) - \frac{1}{2} g(t)^2 \nabla_{\mathbf{x}} \log p_t(\mathbf{x}_t)$. Here, $\nabla_{\mathbf{x}} \cdot v$ denotes the divergence of the velocity field, which can be obtained as the trace of its Jacobian. This term is computationally expensive in high dimensions, which motivates estimating the divergence using the Hutchinson estimator (see Appendix~\ref{appendix:CNFJacobian}).

\subsection{Jacobian determinant estimation}
Continuous flows require Jacobian determinant for likelihood evaluation or importance sampling. Hutchinson’s estimator~\cite{hutchinson1989stochastic} approximates traces via randomized projections,
\begin{equation}
    \hat{T}_{\mathrm{Hutch}} = \mathbf{u}^{\top} 
    \left( \frac{\partial v}{\partial \mathbf{x}} \right) 
    \mathbf{u}, \quad \mathbf{u} \sim \mathcal{D},\; 
    \mathbb{E}[\mathbf{u}\mathbf{u}^\top]=\mathbf{I},
\end{equation}
Here $\mathcal{D}$ is a probe distribution on $\mathbb{R}^D$, such as the standard Gaussian distribution or independent Rademacher entries; $\mathbf{I}$ is the identity matrix. The trace estimate is unbiased, but exponentiation generally biases determinant estimates in high-dimensional flows~\cite{pmlr-v97-behrmann19a,noe2019boltzmann}.

To alleviate the computational cost of evaluating the flow Jacobian at inference, Flow Perturbation (FP)~\cite{peng2025flow} provides an unbiased estimator for the Jacobian determinant when the directional derivative is evaluated exactly:
\begin{equation}
    \hat{\mathcal{J}}_{\mathrm{FP}} = \left\| \frac{\partial \f^{-1}}{\partial \mathbf{x}} \epsilon \right\|^{-D}, \quad
    \epsilon \sim \mathbb{S}^{D-1},
\end{equation}
where $\epsilon$ is sampled uniformly from the unit sphere in $D$ dimensions. For exact inverse derivatives, $\mathbb{E}[\hat{\mathcal{J}}_{\mathrm{FP}}]=|\det(\partial\f/\partial\bz)|$.

\section{Method}
\subsection{Flow Perturbation++ estimator}
\label{sec:method}
\noindent
\begin{minipage}[t]{0.52\linewidth}
\vspace{0pt}
As Figure~\ref{fig:fp_pp_overview} illustrates, numerical integration decomposes the flow $\f$ into $T$ steps,
\begin{equation}
\begin{aligned}
\f &= \f_1 \circ \cdots \circ \f_T,\\
\left|\det\!\left(\frac{\partial \f}{\partial \bz}\right)\right|
&= \prod_{k=1}^T \left|\det(J_k)\right|.
\end{aligned}
\end{equation}
Here $J_k$ is the Jacobian of $\f_k$ at the corresponding point on the trajectory.
FP++ applies a separate perturbation to each inverse step and multiplies the resulting determinant estimates,
\begin{equation}
\hat{\mathcal{J}}_{\mathrm{FP++}}
= \prod_{k=1}^T \|J_k^{-1}\epsilon_k\|^{-D},
\quad \epsilon_k \sim \mathbb{S}^{D-1},
\label{eq:fp_++}
\end{equation}
With independent directions across steps, the product is unbiased for exact inverse-step derivatives:
\begin{equation}
\mathbb{E}[\hat{\mathcal{J}}_{\mathrm{FP++}}]
= \left| \det \frac{\partial \mathbf{f}}{\partial \mathbf{z}} \right|.
\end{equation}
Appendix~\ref{sec:fp_pp} provides a formal proof.

\end{minipage}
\hfill
\begin{minipage}[t]{0.45\linewidth}
\vspace{0pt}
\centering
\includegraphics[width=\linewidth]{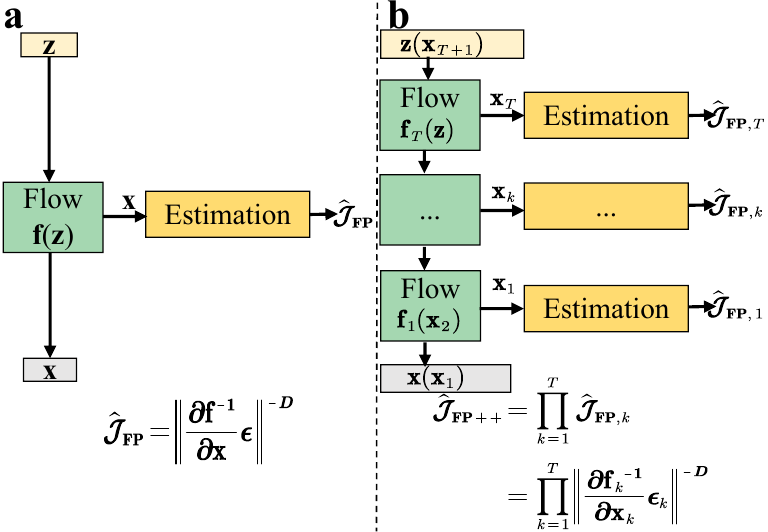}
\captionof{figure}{\textbf{Flow Perturbation vs FP++}.
(a) Single-step FP estimates the full-flow determinant in one shot, which can lead to high variance.
(b) FP++ decomposes flow into $T$ steps and applies determinant estimation sequentially from latent $\bz = \bx_{T+1}$ to $\bx = \bx_1$.
The product of per-step estimates yields the FP++ determinant estimator.}
\label{fig:fp_pp_overview}
\end{minipage}

In practice, the inverse-JVP is approximated via central finite differences:
\begin{equation}
J_k^{-1}\epsilon_k \approx
\frac{\f_k^{-1}(x_k+\delta\epsilon_k)-\f_k^{-1}(x_k-\delta\epsilon_k)}{2\delta}.
\end{equation}
Here $\delta$ is the perturbation size. Both inverse evaluations use the same direction $\epsilon_k$, avoiding explicit construction of $J_k$, while fresh spherical directions are sampled across steps. This finite-difference approximation introduces numerical error and is distinct from the exact-derivative unbiasedness statement above. Convergence and perturbation-size sensitivity are examined in Appendices~\ref{app:convergence} and~\ref{app:fd}, with inverse consistency discussed in Appendix~\ref{app:numerical_inverse_scope}.

\textbf{Variance reduction.} For a fixed trajectory, independent probes eliminate cross-step covariance in the log-domain sum. FP++ has lower log-domain variance when one-step variances match and FP has positive aggregate cross-step covariance. Appendix~\ref{sec:fpplusplus_variance} gives a small-step stretch-alignment condition for the latter; Appendix~\ref{app:direct_variance} evaluates determinant and importance-weight variance.

\subsection{Integration with sequential Monte Carlo}
\label{sec:smc_integration}

\noindent
\begin{minipage}[t]{0.53\linewidth}
\vspace{0pt}
In high-dimensional systems, importance weights often concentrate on only a few samples. We use SMC over $N$ levels, $0=\beta_0<\cdots<\beta_{N-1}=1$, with the ideal exact-Jacobian geometric bridge
\begin{equation}
\pi_n(\bx)\propto p_B(\bx)^{\beta_n}q_f(\bx)^{1-\beta_n},
\end{equation}
and update incremental weights via
\begin{equation*}
\omega_n(\bx)=\omega_{n-1}(\bx)\exp\!\left[-(\beta_n-\beta_{n-1})W(\bx)\right],
\tag{11}
\end{equation*}
where $q_f$ is the flow-induced density and $W(\bx)=\log q_f(\bx)-\log p_B(\bx)$ up to constants. FP++ supplies the determinant correction in $\log q_f$. We resample when $\mathrm{ESS}=(\sum_j(\omega_n^{(j)})^2)^{-1}$ falls below a threshold.
\end{minipage}
\hfill
\begin{minipage}[t]{0.45\linewidth}
\vspace{0pt}
\refstepcounter{algorithm}
\label{alg:smc_fp_pp}
\hrule
\vspace{1pt}
\noindent{\small\textbf{Algorithm \thealgorithm}\quad SMC with Flow Perturbation++}
\vspace{1pt}
\hrule
\vspace{2pt}
\small
\textbf{Input:} $q_f,p_B,M,\{\beta_n\}_{n=0}^{N-1},\tau,L$\\[-1pt]
\textbf{Init:} $\bx_0^{(i)}\sim q_f,\ \omega_0^{(i)}=1/M$\\[-1pt]
\textbf{for } $n=1$ to $N-1$:
\begin{enumerate}
\item Compute the FP++ determinant correction entering $W^{(i)}$
\item Update and normalize weights:
{
\[
\omega_n^{(i)} \propto \omega_{n-1}^{(i)}e^{-(\beta_n-\beta_{n-1})W^{(i)}}
\]}
\item If $\mathrm{ESS}_n<\tau M$, resample and set $\omega_n^{(i)}=1/M$
\item Apply $L$ $\pi_{\beta_n}$-invariant $K_{\beta_n}$ moves
\end{enumerate}
\textbf{Output:} $\{\bx_{N-1}^{(i)},\omega_{N-1}^{(i)}\}$
\vspace{2pt}
\hrule
\end{minipage}

The full FP++--SMC procedure is summarized in Algorithm~\ref{alg:smc_fp_pp}. With stochastic determinant estimates, the intermediate targets are extended-state targets (Appendix~\ref{app:smc_protocol}), not this exact-Jacobian bridge; the ideal endpoint marginal remains the Boltzmann target. Finite-particle and numerical approximations remain.

\subsection{Computational complexity comparison}
\label{sec:method_Cost}

We compare the computational cost of baseline Jacobian-related estimators with the proposed FP++ (multi-step) approach. 
Baselines include single-step Flow Perturbation (FP)~\cite{peng2025flow}, the Hutchinson trace estimator~\cite{hutchinson1989stochastic}, and the exact brute-force Jacobian (BFJacob).

Table~\ref{table:compare} summarizes their computational characteristics. All methods require one forward ODE integration to generate the trajectory $\bx = f(\bz)$; additional cost arises from estimating the Jacobian-related determinant correction.

\paragraph{Cost unit.}
We measure computational complexity in terms of \emph{ODE-equivalent passes}, where one pass corresponds to a full trajectory integration (forward or backward along the flow).
\begin{itemize}
\item \textbf{BFJacob:} Computes the full $D \times D$ Jacobian via $D$ vector–Jacobian products (VJPs). Each VJP requires one backward pass along the flow. The total cost is therefore $1 + D$ passes.
\item \textbf{Hutchinson estimator:} Approximates the trace using $N$ random probe vectors. Each probe requires one VJP, resulting in a total cost of $1 + N$ passes.
\item \textbf{FP / FP++:} We estimate the Jacobian-related term using finite-difference evaluations of the inverse flow. Each probe requires two evaluations of the inverse dynamics along the trajectory. Therefore, using $N$ probes leads to a total cost of $1 + 2N$ passes.
\end{itemize}

\noindent
\begin{minipage}[t]{0.59\linewidth}
\vspace{0pt}
\centering
\captionof{table}{Computational cost comparison of Jacobian-related estimators. Cost is measured in ODE-equivalent passes. $N$ denotes the number of probes; in our experiments, we set $N=1$ for FP / FP++.}
\label{table:compare}
\small
\begin{tabular}{@{}lccc@{}}
\toprule
Method & Jacobian Calc. & Biased & Cost (\# passes) \\
\midrule
FP / FP++ & No & No & $1+2N$ \\
BFJacob & Yes & No & $1+D$ \\
Hutchinson & No & Yes & $1+N$ \\
\bottomrule
\end{tabular}
\end{minipage}
\hfill
\begin{minipage}[t]{0.38\linewidth}
\vspace{0pt}
\centering
\includegraphics[width=0.866\linewidth]{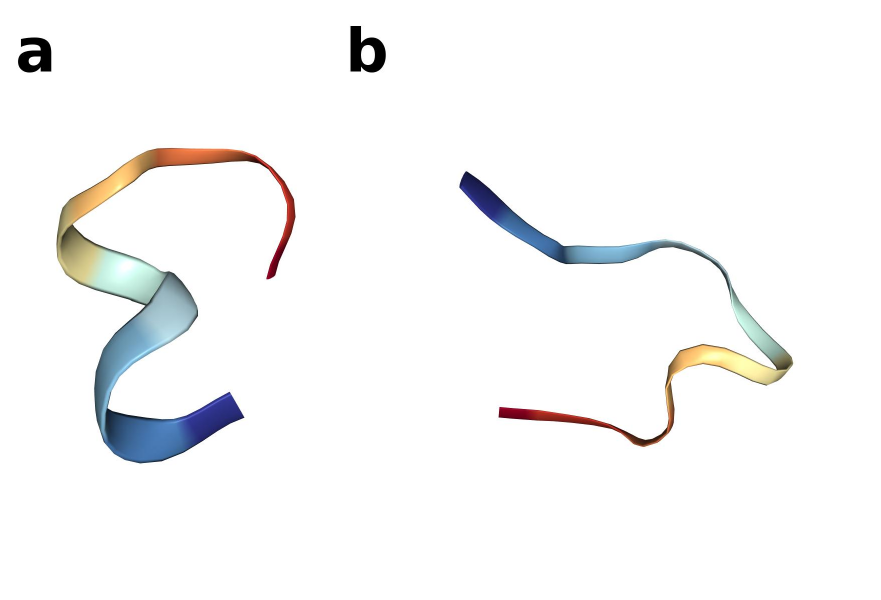}
\captionof{figure}{Representative CNL025 configurations at 300\,K: (a) partially $\alpha$-helical; (b) folded $\beta$-hairpin.}
\label{fig:chignolin_structures}
\end{minipage}

\section{Experiments and results}
\label{sec:experiments}
We evaluate the FP++-SMC framework on two representative high-dimensional systems: a synthetic Gaussian Mixture Model (GMM), and the all-atom Chignolin peptide.

\subsection{Experimental setup}
\label{sec:setup_metrics}

\subsubsection{Test systems}
\paragraph{Gaussian mixture model (GMM).}
We consider a high-dimensional two-component anisotropic Gaussian mixture as a benchmark system. Training samples are drawn uniformly from the mixture, while the target Boltzmann distribution assigns asymmetric component weights $(0.25,\,0.75)$, making it challenging for models to fully match the target distribution. This bimodal system is structurally simple, yet poses a nontrivial sampling challenge due to metastability between modes, and admits an analytically tractable reference solution, making it an ideal platform for evaluating the performance of FP++. Full system specifications are provided in Appendix~\ref{sec:GMM}.

\paragraph{Chignolin mutant CNL025.}
The 10-residue Chignolin mutant CNL025~\cite{Rodriguez:2011uo, McKiernan:2017wh, Maruyama18} provides a compact and challenging biomolecular test case. 
We adopt the CHARMM22 force field~\cite{MacKerell-Jr.:2000tn} with the OBC2 implicit solvent model~\cite{Onufriev:2004vx}. 
Training configurations at 300\,K are extracted from replica-exchange molecular dynamics data released by the No\'e group.\footnote{\url{http://ftp.mi.fu-berlin.de/pub/cmbdata/bgmol/datasets/chignolin/ChignolinOBC2PT.tgz}}
Representative conformations are shown in Fig.~\ref{fig:chignolin_structures}.

\subsubsection{Flow models}

For all experiments, we use the pretrained continuous-time flow models released in~\cite{peng2025flow} without further training.
The models were trained with the $v$-prediction objective~\cite{salimans2022progressive}, which improves numerical stability in high dimensions, with the neural network representing the time-dependent velocity field $v_\theta(\mathbf{x}, t)$ that defines the probability flow.
Full architecture and training details are reported in~\cite{peng2025flow}.

For the GMM benchmark, a lightweight multilayer perceptron is sufficient to model the low-complexity vector field associated with the mixture structure.
For Chignolin, we employ a Transformer-based architecture~\cite{vaswani2017attention,li2022dit} to parameterize $v_\theta(\mathbf{x}, t)$, enabling the model to capture long-range geometric and many-body correlations in high-dimensional biomolecular configurations.

\subsubsection{Compared Jacobian estimators}

We evaluate estimators under identical SMC pipelines, all using a 20-step RK4 ODE solver:
\begin{itemize}
    \item \textbf{FP / FP++ (single-probe, $N=1$)}~\cite{peng2025flow}, perturbation-based estimators using one finite-difference probe.
    \item \textbf{Hutchinson trace estimators}~\cite{hutchinson1989stochastic}, using Gaussian or Rademacher probes, denoted as ‘G$n$’ and ‘R$n$’ respectively, with $n\in\{1,2,10\}$.
    \item \textbf{Brute-force Jacobian (BFJacob)}, an exact but computationally expensive method via automatic differentiation.

\end{itemize}

\subsubsection{Evaluation metrics}

We report several standard metrics to quantify estimator accuracy and sampling performance.
\begin{itemize}
\item \textbf{Energy Distribution.}
We compare empirical SMC energy histograms with the reference Boltzmann distribution.

\item \textbf{Computational Cost.}
We measure the approximate wall-clock time required for one full SMC run.

\item \textbf{Reaction-Coordinate Occupancy.}
We compute occupancy along a selected reaction coordinate to evaluate recovery of the free-energy landscape.

\item \textbf{Remaining Ancestor Count.}
We track the number of surviving ancestors after resampling~\cite{del2006sequential}, a diagnostic indicator of particle degeneracy. 

\end{itemize}
In addition, we include supplementary evaluations in Appendix~\ref{appendix:extra_results}, including variance analysis of the log-domain Jacobian estimator across methods and dimensionalities, convergence studies of FP++ and Hutchinson-based estimators, stability evaluations on Chignolin, sensitivity analyses with respect to finite-difference step size, direct determinant/weight variance and ESS measurements, and an additional validation on a molecular system, alanine dipeptide from Transferable Boltzmann Generators (TBG)~\cite{klein2024transferable}.
\subsection{Results}
\label{sec:results}
Before presenting the specific tasks, we briefly discuss the numerical robustness of the proposed FP++ estimator. We conduct sensitivity and stability analyses (see Appendix~\ref{appendix:extra_results}) to assess its reliability in importance sampling. Empirically, FP++ achieves a variance level comparable to the brute-force Jacobian even with a single probe, and remains stable across dimensionalities ranging from 10D to 1000D. Our sensitivity study further shows that setting the finite-difference step size $\delta = 0.001$ yields accurate and stable estimates consistent with the JVP reference across all tested systems. In addition, repeated independent runs show that the estimator mean converges to the brute-force Jacobian reference as the number of runs $M$ increases. These results confirm that FP++ provides a stable and low-variance estimator suitable for the subsequent Boltzmann sampling experiments.
\subsubsection{Low-dimensional GMM (10D)}

\begin{figure}[t]
\centering

\begin{minipage}{0.4\linewidth}
    \centering
    \includegraphics[width=\linewidth]{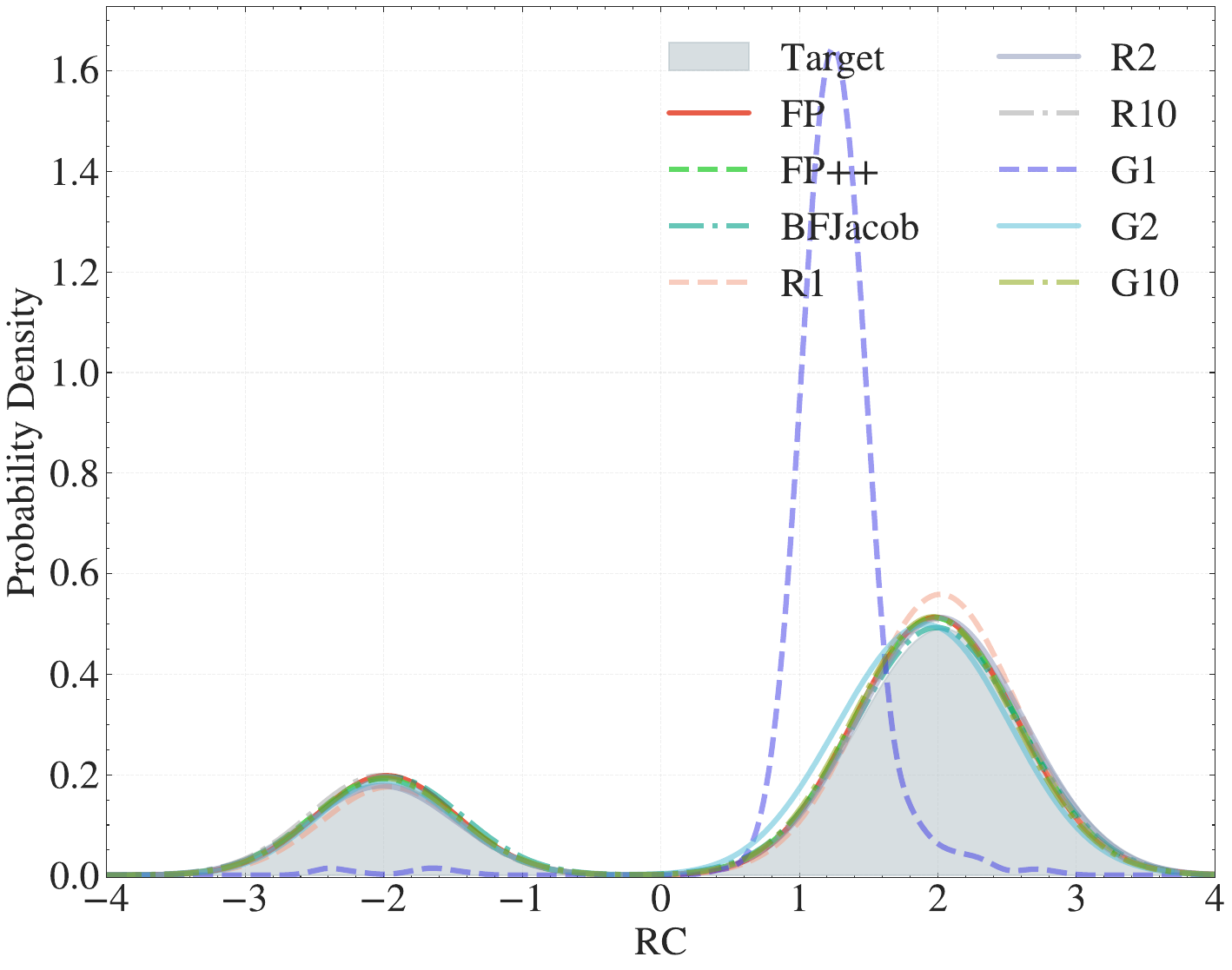}
    \caption{
    Final SMC sample distribution along the reaction coordinate for the 10D GMM. 
    The solid blue curve shows the target distribution obtained via direct sampling.
    }
    \label{fig:GMM10d}
\end{minipage}
\hfill
\begin{minipage}{0.58\linewidth}
    \centering
    \captionof{table}{
    Performance of baseline and proposed estimators on the 10D GMM.  
    The modal weight reports the fraction of SMC particles located in the first Gaussian component.
    }
    \label{table:GMM10d_results}
    \small
    \begin{tabular}{l p{0.9cm} p{0.9cm} p{0.8cm} p{0.9cm}}
    \toprule
    Estimator & Mean & Std & \# Preserved & Time \\
    \midrule
    Hutchinson G1 & 0.0189 & 0.018 & 131 & 2.5m  \\
    Hutchinson G2 & 0.2278 & 0.013 & 828 & 3.0m \\
    Hutchinson G10 & 0.2454 & 0.011 & 943 & 3.0m \\
    Hutchinson R1 & 0.2289 & 0.017 & 972 & 2.5m \\
    Hutchinson R2 & 0.2391 & 0.010 & 971 & 3.0m \\
    Hutchinson R10 & 0.2503 & 0.016 & 971 & 3.0m \\
    BF Jacobian & 0.2497 & 0.019 & 978 & 5.5m \\
    FP (single-step) & 0.2499 & 0.015 & 862 & 2.5m \\
    \textbf{FP++ (multi-step)} & 0.2503 & 0.011 & 957 & 2.5m \\
    \bottomrule
    \end{tabular}
\end{minipage}

\end{figure}

We assess estimator performance on a 10-dimensional Gaussian Mixture Model using SMC with 1{,}000 particles across 10 independent runs.  
Accuracy is measured by the fraction of particles assigned to the first Gaussian component.  
As shown in Table~\ref{table:GMM10d_results}, both the FP baseline and FP++ accurately recover the target modal probability (25\%), matching the ground truth.  
Hutchinson estimators exhibit noticeable bias when using only a few random probe vectors ($n=1,2$), with accuracy improving as the number of probe noises increases.  
The brute-force Jacobian (``BF Jacobian''), using results reported in~\cite{peng2025flow}, provides high accuracy but at substantially greater computational cost.
The reaction-coordinate distributions in Fig.~\ref{fig:GMM10d} further demonstrate that FP++ yields stable and accurate reconstruction of the target distribution.
\subsubsection{1000 dimensional GMM}
\begin{figure*}[!htb]
    \centering
    \includegraphics[width=0.75\linewidth]{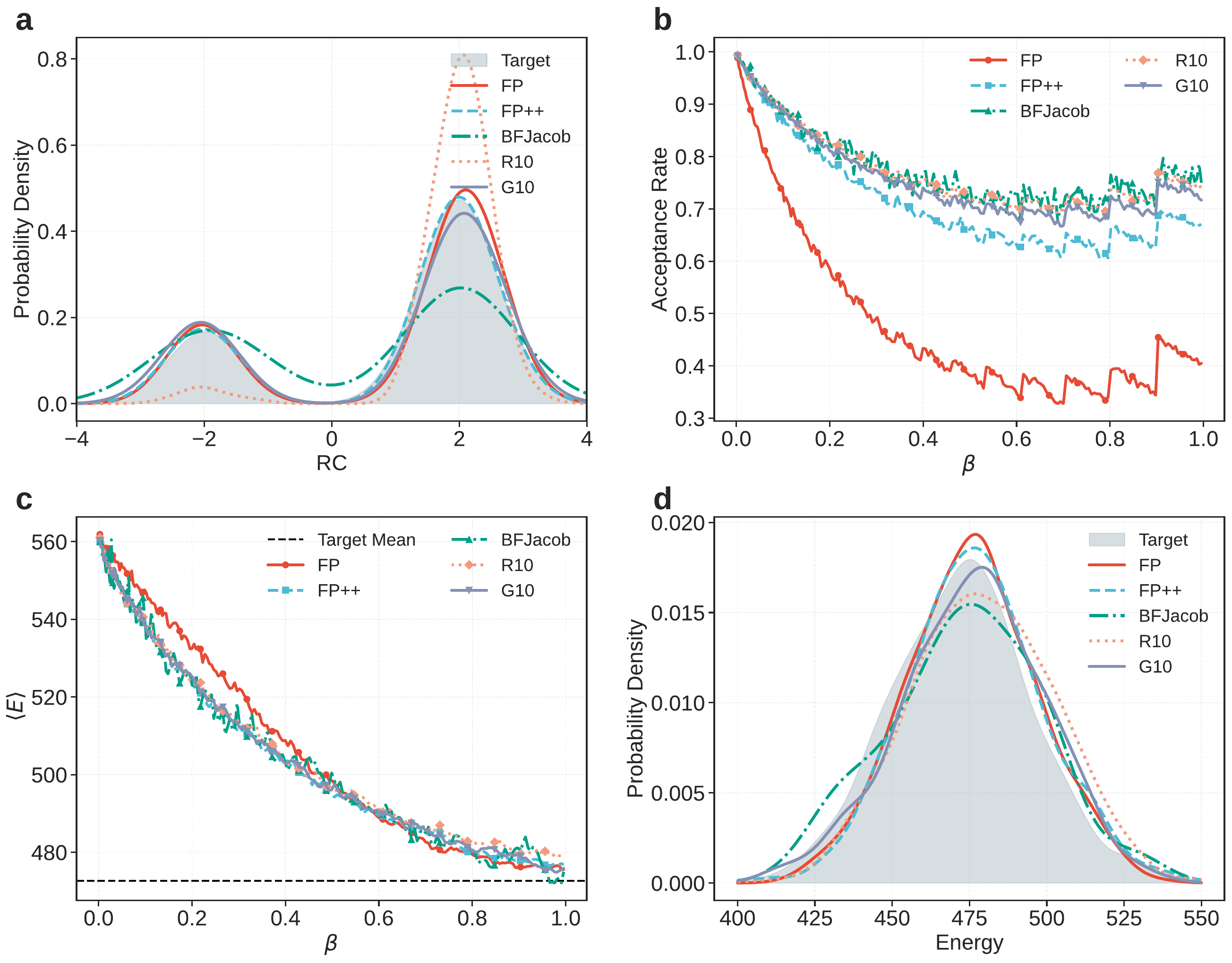}
\caption{
SMC results for the 1000-dimensional GMM. 
(a) Probability distribution of final SMC samples along the reaction coordinate. 
(b) Monte Carlo acceptance rate as a function of \(\beta\). 
(c) Average sample energy as a function of \(\beta\). 
(d) Final energy distribution. The target energy and distribution (solid line) are obtained by direct sampling from the target GMM. 
All results are derived from a single SMC run.
}

    \label{fig:GMM1000d}
\end{figure*}
We further benchmark all estimators on a substantially more challenging 1000-dimensional GMM.
For FP, FP++, and Hutchinson variants, SMC is performed with 1{,}000 particles and ten independent runs, using 2{,}000 annealing steps with 10 MCMC transitions per step.
Due to its prohibitive cost, the Brute-Force Jacobian baseline (BFJacob) is evaluated using 100 particles in a single run, using results reported in~\cite{peng2025flow}.

\begin{table}[t]
\centering

\begin{minipage}[t]{0.49\linewidth}
\centering
\caption{Performance of Jacobian estimators on the 1000-dimensional GMM.
Modal weights report the fraction of particles assigned to the first Gaussian component.}
\label{tab:compare_GMM}

\small
\setlength{\tabcolsep}{3pt}
\begin{tabular}{@{}lcccc@{}}
\toprule
Estimator & Mean & Std & \# Preserved & Time \\
\midrule
G1 & 0.534 & 0.073 & 154 & 0.7 h \\
G2 & 0.377 & 0.023 & 222 & 1.1 h \\
G10 & 0.277 & 0.022 & 360 & 5.1 h \\
R1 & 0 & 0 & 328 & 0.7 h \\
R2 & 0 & 0 & 361 & 1.1 h \\
R10 & 0.044 & 0.011 & 409 & 5.1 h \\
BF Jacobian & 0.400 & -- & 45 & 31 h \\
FP & 0.209 & 0.083 & 59 & 1.1 h \\
\textbf{FP++} & \textbf{0.256} & 0.027 & 312 & 1.1 h \\
\bottomrule
\end{tabular}

\end{minipage}
\hfill
\begin{minipage}[t]{0.49\linewidth}
\centering
\captionof{table}{Performance of different Jacobian estimators on the Chignolin mutant.
The table reports the fraction of particles in the $\beta$-hairpin basin (Fig.~\ref{fig:CGN}), along with preserved lineages and computational cost.
Mean, Std, and preserved lineages summarize ten runs; time is per run.
Target value is 0.385; base flow yields 0.311.}
\label{table:compare_GCN}

\small
\setlength{\tabcolsep}{1.5pt}
\begin{tabular}{@{}lcccc@{}}
\toprule
Method & Mean & Std & \# Preserved & Time \\
\midrule
R2 & 0.2330 & 0.0168 & 1529 & 30.4 h \\
R10 & 0.2457 & 0.0369 & 1541 & 100 h \\
G2 & 0.2136 & 0.0685 & 1316 & 30.4 h \\
G10 & 0.2303 & 0.0394 & 1458 & 100 h \\
FP & 0.4450 & 0.0510 & 1433 & 11.1 h \\
\textbf{FP++} & \textbf{0.3658} & \textbf{0.0146} & 1474 & 11.1 h \\
\bottomrule
\end{tabular}

\end{minipage}

\end{table}

Table~\ref{tab:compare_GMM} reports the estimated probability of the first mode (ground truth $25\%$), lineage preservation, and computational cost.
Among all methods, \textbf{FP++ best recovers the target modal probability}, yielding $0.256 \pm 0.027$ across ten runs while preserving a large number of lineages (312 on average).
Notably, FP++ achieves this accuracy at essentially the same computational cost as the single-step FP estimator.

Using 10 Gaussian Hutchinson probes (G10) attains the next-best performance ($0.277 \pm 0.022$), but requires roughly five times the computational cost.
The single-step FP estimator produces reasonable estimates ($0.209 \pm 0.083$) with substantially higher variance.
All other Hutchinson variants suffer from significant bias or mode collapse, accompanied by severe lineage degeneracy.
BFJacob underperforms at this scale due to the limited particle budget; extrapolating to 1{,}000 particles would increase its computational cost by roughly 300× compared with FP++, making it entirely impractical.

Figure~\ref{fig:GMM1000d} shows final sample and energy distributions (a,d) and acceptance rates and average energies across the annealing schedule (b,c).
FP++ consistently matches or exceeds the accuracy of both G10 and BFJacob while using only a fraction of their computational budget.
Partial coordinate updates induce stepwise acceptance-rate fluctuations, with FP exhibiting the highest variance and lowest acceptance rates.
FP++ reduces this variance by per-step Jacobian decomposition, resulting in consistently higher acceptance rates.

\subsubsection{Chignolin mutant}

For the Chignolin mutant, we conducted SMC simulations using the FP, FP++, and Hutchinson estimators.
Each simulation employed 2{,}000 particles, 600 intermediate distributions, and 2 MCMC steps per level.
Resampling was performed using the KL Reshuffling scheme~\cite{kviman2022statistical}.
The Brute-Force Jacobian baseline (BFJacob) was omitted due to its prohibitive computational cost: even a single run with a reduced number of particles would already require orders of magnitude more resources than FP++ or Hutchinson estimators, making it infeasible for this system.  
\begin{figure}[!t]
    \centering
    \includegraphics[width=0.80\linewidth]{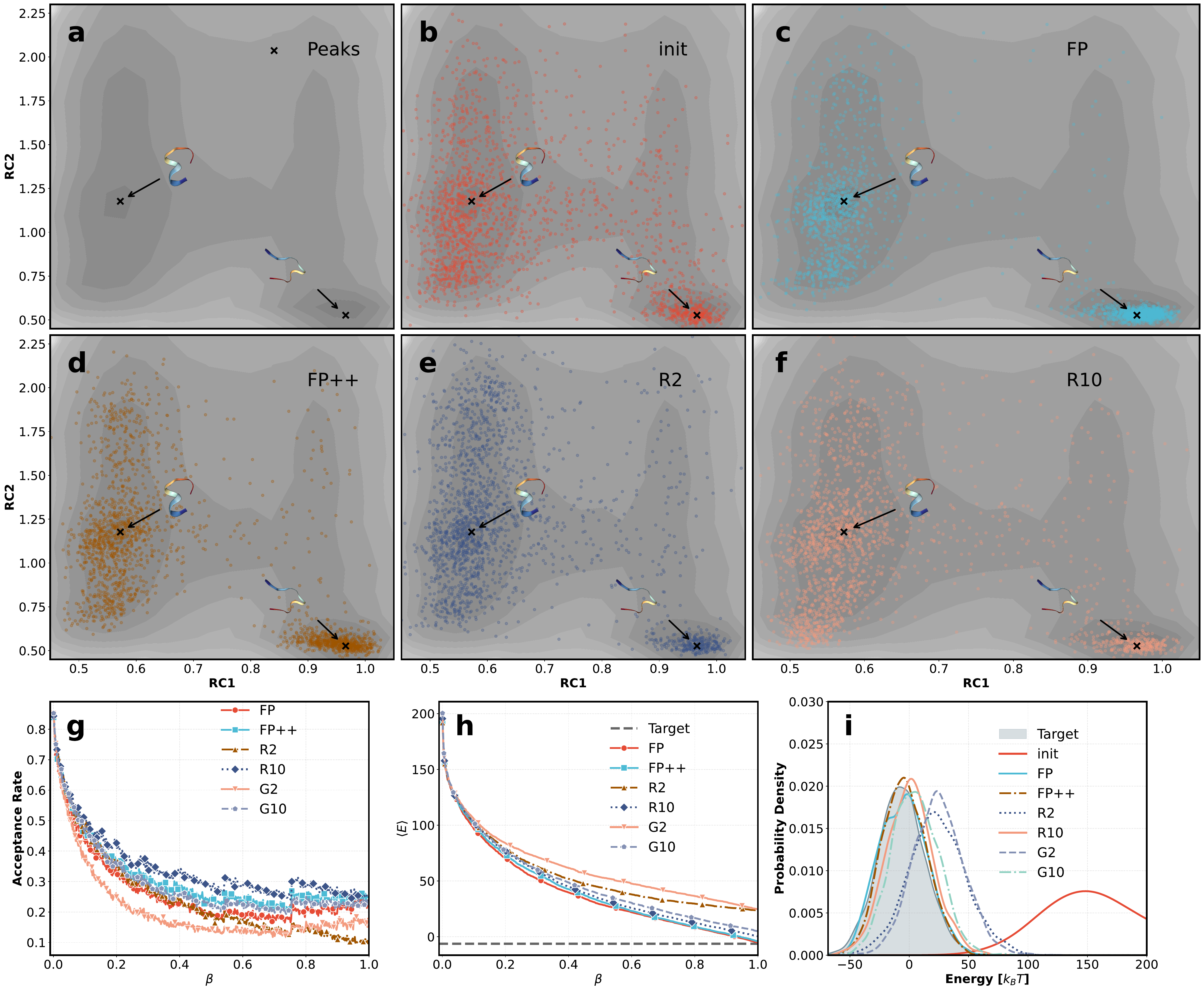}
    \caption{SMC results for the Chignolin mutant.
(a--f) Sample distributions projected onto the reaction-coordinate (RC) plane:
(a) Target distribution derived from the training data;
(b) Samples generated directly by the base flow model;
(c) Samples produced by SMC using the single-step FP estimator;
(d) Samples produced by SMC using the FP++ estimator;
(e) Samples produced by SMC with the Hutchinson estimator (10 Rademacher vectors);
(f) Samples produced by SMC with the Hutchinson estimator (10 Gaussian vectors).
(g) Monte Carlo acceptance rates as functions of $\beta$;
(h) Average energy $\langle E \rangle$ of intermediate SMC samples;
(i) Energy distributions at the end of SMC compared to the target distribution and the base flow (``init'').
All SMC results are derived from a single run.}
    \label{fig:CGN}
\end{figure}

Table~\ref{table:compare_GCN} reports the fraction of particles occupying the right (folded) basin, with a target occupancy of $0.385$ and a base-flow value of $0.311$.
Among all tested estimators, \textbf{FP++ achieves the closest agreement with the target}, yielding $0.3658\pm0.0146$ across ten runs (range $0.3455$--$0.3875$) while preserving 1{,}474 particles on average.
The single-step FP estimator overshoots the target at $0.4450\pm0.0510$ (ten-run range $0.3320$--$0.4830$) under the same computational budget.
Hutchinson-based estimators systematically underestimate basin occupancy, even with 10 Gaussian or Rademacher probes.
Variants with fewer probes (G2/R2) remain less accurate than FP++ under this matched SMC setting; complete ten-run statistics and ranges are reported in Appendix~\ref{appendix:Chignolin}, Table~\ref{tab:chignolin_stability_compact}.

Although finite-difference approximations incur a comparable number of model evaluations to low-probe Hutchinson estimators (e.g., G2/R2), they can be substantially faster in practice for Transformer-based flows.
As shown in Table~\ref{table:compare_GCN}, finite-difference--based FP and FP++ achieve significantly lower wall-clock times than Hutchinson-type estimators under comparable or even larger computational budgets. This behavior arises from the use of self-attention and normalization layers:
Jacobian--vector products require backpropagation through these components,
whose computational cost is dominated by dense attention operations and softmax derivatives.
Such backward passes are well known to be considerably more expensive than forward evaluations~\cite{vaswani2017attention,liang2024conv}. In contrast, finite-difference estimators only involve multiple forward passes, which are heavily optimized on modern GPUs via fused kernels and specialized attention implementations. For consistency, Appendix~\ref{app:jvp_cost} reports runtimes with all methods implemented using VJP or JVP, including FP and FP++.

Figure~\ref{fig:CGN} illustrates the projected particle distributions along the reaction-coordinate plane.
Both FP and FP++ faithfully recover the bimodal target distribution, with negligible leakage into the low-probability corridor separating the $\alpha$-helical and $\beta$-hairpin basins.
In contrast, Hutchinson-based and base-flow samples display substantial spurious occupancy between basins.

Energy histograms (Figure~\ref{fig:CGN}(i)) further confirm that FP and FP++ generate high-fidelity ensembles compared to Hutchinson estimators or direct base-flow samples.
Acceptance rates along the annealing schedule (Figure~\ref{fig:CGN}(g)) remain smooth, consistent with largely fixed-coordinate updates, and FP++ exhibits a slight improvement over FP, reflecting reduced variance through the multi-step decomposition.

Complementary diagnostics for representative Chignolin runs with 2,000 particles are reported in Appendix~\ref{appendix:Chignolin} (Table~\ref{tab:chignolin_metrics}). FP++ has the highest pre-resampling importance ESS/$N$ ($0.003499$) and a torsion-W2 ($2.7349$) close to R10 ($2.6891$); FP has lower Energy-W2 ($1.6512$ versus $2.7330$). R2/R10 retain more ancestry but underestimate folded-basin occupancy. These weight, distributional, and genealogical metrics reveal trade-offs rather than uniform improvement.

For reference, conventional MD trajectories often become trapped in single metastable states, underscoring the challenges of effectively sampling the Chignolin mutant using standard simulation approaches (Appendix~\ref{appendix:MD}).

\section{Conclusion}
We have introduced \textbf{Flow Perturbation++ (FP++)}, a theoretically grounded estimator of the flow Jacobian determinant that leverages the semigroup property of flow maps to decompose the global Jacobian into per-step determinants. It is determinant-unbiased for exact inverse-step derivatives, with variance reduction under the conditions in Appendix~\ref{sec:fpplusplus_variance}. FP++ serves as a practical and scalable alternative to Hutchinson-style trace estimators~\cite{hutchinson1989stochastic}.
When integrated with sequential Monte Carlo (SMC), FP++ enables accurate Boltzmann reweighting and reliable thermodynamic statistics across both synthetic high-dimensional Gaussian mixtures and realistic biomolecular systems such as the Chignolin mutant.
Our experiments demonstrate that FP++ not only improves estimator accuracy and stability compared to both the original FP and Hutchinson estimators, but also substantially reduces computational cost relative to conventional brute-force Jacobian evaluation.

\paragraph{Limitations.}
Finite differences and reverse integration introduce numerical errors (Appendices~\ref{app:fd} and~\ref{app:numerical_inverse_scope}), and variance reduction is conditional rather than universal. Exact Jacobians remain preferable when numerical approximations cannot be validated. FP++ complements sampler pipelines; end-to-end comparisons with FALCON or related samplers remain future work.

\begin{ack}
This work was supported by the Opening Project of the State Key Laboratory of Information Photonics and Optical Communications (Grant No. IPOC2025ZT02), the National Natural Science Foundation of China (Grant No. 22473016), the Beijing Natural Science Foundation (Grant No. L258008), and the Fundamental Research Funds for Beijing University of Posts and Telecommunications (Grant No. 2025AI4S01). We are thankful to the anonymous reviewers for their helpful comments. We thank the Shanghai Academy of AI for Science and Fudan University for providing computational resources through the CFFF platform during the rebuttal period. The authors declare no competing interests.
\end{ack}\ \textbf{Software and Data.}
\label{sec:software_data}
Implementations of the methods and scripts for data generation and evaluation are publicly available at \url{https://github.com/XinPeng76/Flow_Perturbation_pp}.

\clearpage
\bibliographystyle{plainnat}
\small
\bibliography{references}
\normalsize

%%%%%%%%%%%%%%%%%%%%%%%%%%%%%%%%%%%%%%%%%%%%%%%%%%%%%%%%%%%%

\newpage
\appendix
\section{Detailed sequential Monte Carlo protocol}
\label{app:smc_protocol}

This section describes the SMC procedure used with the different Jacobian estimators. The particle count, annealing schedule, mutation kernels, and resampling rule are matched across methods. For stochastic determinant estimators, the target distribution is formulated on an extended state space that includes the probes.

\paragraph{Intermediate targets and weights.}
Let $q(\bz)$ be the latent density, $\bx=\f(\bz)$, and $p_B(\bx)\propto\exp[-E(\bx)]$. With an exact Jacobian, the usual geometric bridge in configuration space is $\pi_\beta(\bx)\propto q_f(\bx)^{1-\beta}p_B(\bx)^\beta$. For FP++, let $\xi$ denote the complete collection of raw auxiliary variables and let $\rho(\xi)$ be their joint density. Each spherical probe is represented as $\epsilon=\xi/\|\xi\|$ with $\xi\sim\mathcal N(0,I)$, so refreshing Gaussian coordinates preserves the uniform spherical direction after normalization. We write $\widehat{\mathcal J}(\bz,\xi)$ for the resulting positive determinant estimator and define
\begin{equation}
\widehat{\mathcal J}(\bz,\xi)=\exp[\Delta S(\bz,\xi)],
\qquad
R(\bz,\xi)=\frac{p_B(\f(\bz))}{q(\bz)}\widehat{\mathcal J}(\bz,\xi).
\label{eq:smc_extended_ratio}
\end{equation}
The corresponding extended targets are
\begin{equation}
\widetilde\pi_\beta(\bz,\xi)
\propto q(\bz)\rho(\xi)R(\bz,\xi)^\beta.
\label{eq:smc_bridge_appendix}
\end{equation}
At $\beta=0$, Eq.~\eqref{eq:smc_bridge_appendix} is the latent proposal augmented by independent probes. For $0<\beta<1$, its configuration marginal generally differs from $q_f^{1-\beta}p_B^\beta$ because $\mathbb E[\widehat{\mathcal J}^{\,\beta}]\ne\mathbb E[\widehat{\mathcal J}]^\beta$ in general. The endpoint marginal is characterized by the following proposition.

\begin{proposition}[Endpoint marginal of the ideal extended target]
\label{prop:extended_smc_endpoint}
Assume that $\f$ is a diffeomorphism, $\widehat{\mathcal J}(\bz,\xi)>0$, and
\begin{equation}
\mathbb E_{\xi\sim\rho}\!\left[\widehat{\mathcal J}(\bz,\xi)\right]
=\left|\det D\f(\bz)\right|
\quad\text{for }q\text{-almost every }\bz.
\label{eq:extended_unbiased_condition}
\end{equation}
If $0<Z_B=\int p_B(\bx)\,d\bx<\infty$, then the $\bz$-marginal of the normalized target $\widetilde\pi_1$ is
\begin{equation}
\widetilde\pi_1(\bz)
=\frac{p_B(\f(\bz))\left|\det D\f(\bz)\right|}{Z_B},
\end{equation}
and its pushforward through $\f$ is $p_B(\bx)/Z_B$.
\end{proposition}

\begin{proof}
At $\beta=1$, Eqs.~\eqref{eq:smc_extended_ratio}--\eqref{eq:smc_bridge_appendix} give the unnormalized joint density $\rho(\xi)p_B(\f(\bz))\widehat{\mathcal J}(\bz,\xi)$. Integrating over $\xi$ and applying Eq.~\eqref{eq:extended_unbiased_condition} yields $p_B(\f(\bz))|\det D\f(\bz)|$. Its integral over $\bz$ equals $Z_B$ by change of variables, and the same change of variables proves the pushforward statement.
\end{proof}

The proposition characterizes the ideal extended target; the finite-$M$ particle population provides an SMC approximation to it. Each particle stores the complete pair $(\bz,\xi)$. Initialization draws $(\bz,\xi)\sim q\rho$, resampling copies the complete pair, and probe variables are updated through a mutation kernel that leaves $\widetilde\pi_{\beta_n}$ invariant. Independent regeneration of the entire probe collection after each incremental-weight update would change the algorithm. Algorithm~\ref{alg:smc_fp_pp} omits this auxiliary-state bookkeeping for readability.

For a particle carried from level $n-1$, the incremental log weight is
\begin{equation}
\Delta\log\omega_n
=(\beta_n-\beta_{n-1})\log R(\bz,\xi)
=-(\beta_n-\beta_{n-1})W(\bz,\xi),
\label{eq:smc_increment_appendix}
\end{equation}
up to constants common to all particles, where $W=u_X(\f(\bz))-u_Z(\bz)-\Delta S(\bz,\xi)$. The implementation stores Eq.~\eqref{eq:smc_increment_appendix} in the log domain for numerical stability. Determinant-level unbiasedness applies to the positive factor $\exp(\Delta S)$ under Eq.~\eqref{eq:extended_unbiased_condition}; Appendices~\ref{app:fd} and~\ref{app:numerical_inverse_scope} characterize the finite-difference and solver-inversion approximations used in practice.

\paragraph{Normalization, resampling, and mutation.}
At each level, we add Eq.~\eqref{eq:smc_increment_appendix} to the accumulated log weights and normalize them using a log-sum-exp operation, producing $\bar\omega_n^{(i)}$. We then compute
\begin{equation}
\mathrm{ESS}_n=\left(\sum_{i=1}^{M}\bigl(\bar\omega_n^{(i)}\bigr)^2\right)^{-1}.
\end{equation}
The implementation uses $\tau=0.95$: resampling is triggered when $\mathrm{ESS}_n<0.95M$, with an additional near-endpoint resampling in the reported runs. GMM experiments use systematic resampling, whereas Chignolin uses KL reshuffling~\cite{kviman2022statistical}; the weights are reset to $1/M$ after either operation, and final observables use the weighted or resampled population specified with each result.

The annealing values are generated identically for both systems. For $j=0,\ldots,N-1$, the code evaluates
\begin{equation}
\beta_j=
\frac{\sigma\!\left(-2+3j/(N-1)\right)-\sigma(-2)}
{\sigma(1)-\sigma(-2)},
\qquad \sigma(a)=\frac{1}{1+e^{-a}},
\label{eq:implemented_beta_schedule}
\end{equation}
which places more intermediate distributions near the ends than a linear schedule.

Mutation uses a coordinate-refresh Metropolis kernel. Let $s=(\bz,\xi)$ and let $Q_0(s,s')$ denote the joint coordinate-refresh proposal density. For a general proposal, the Metropolis--Hastings acceptance probability targeting Eq.~\eqref{eq:smc_bridge_appendix} is
\begin{equation}
\alpha_\beta(s,s')
=1\wedge
\frac{q(\bz')\rho(\xi')R(\bz',\xi')^\beta Q_0(s',s)}
{q(\bz)\rho(\xi)R(\bz,\xi)^\beta Q_0(s,s')}.
\label{eq:general_extended_mh}
\end{equation}
When $Q_0$ is reversible with respect to the base law $q\rho$, the base and proposal factors cancel, giving $1\wedge\exp\{\beta(\log R'-\log R)\}$. The reported coordinate-refresh implementation satisfies this condition by drawing refreshed latent coordinates from their factorized Gaussian base marginals and raw auxiliary coordinates from $\mathcal N(0,1)$. The latent refresh standard deviations are $t_{\max}=1.0$ for GMM and $0.75$ for Chignolin; the auxiliary variables have unit standard deviation. At level $j$, the kernel refreshes random subsets of $K_x(j)$ latent coordinates and $K_\epsilon(j)$ raw auxiliary coordinates. With $b_j=(j/(N-1))^2$, the integer-valued schedules are
\begin{equation}
K_s(j)=\left\lfloor K_{s,\min}b_j^{c_s}
+K_{s,\max}\bigl(1-b_j^{c_s}\bigr)\right\rfloor,
\qquad s\in\{x,\epsilon\}.
\label{eq:mutation_coordinate_schedule}
\end{equation}
Thus the number of refreshed coordinates decreases over annealing. Equation~\eqref{eq:general_extended_mh} satisfies detailed balance for $\widetilde\pi_{\beta_j}$, and base reversibility yields the simplified acceptance rule used in the experiments. All schedules, proposals, and acceptance rules are held fixed across Jacobian-estimator variants within a system.

\paragraph{Diagnostics and reported configurations.}
Importance ESS and weight-variance diagnostics are computed from the normalized weights immediately before resampling. After resampling, the uniform weights make importance ESS equal to the particle count, so particle ancestry is used to assess diversity: lineage (preservation) ESS and the number or fraction of uniquely preserved ancestors quantify genealogical degeneracy. Final molecular observables are evaluated on the resulting weighted or resampled particle population, as specified with each result. Table~\ref{tab:smc_configurations} lists the matched configurations.

\begin{table}[h]
\centering
\caption{Matched SMC configurations used in the high-dimensional experiments. Here $(K_{x,\max}\!\to\!K_{x,\min};c_x)$ and $(K_{\epsilon,\max}\!\to\!K_{\epsilon,\min};c_\epsilon)$ parameterize Eq.~\eqref{eq:mutation_coordinate_schedule}.}
\label{tab:smc_configurations}
\footnotesize
\setlength{\tabcolsep}{3.2pt}
\begin{tabular}{@{}lrrrrlll@{}}
\toprule
System & $M$ & $N$ & $L$ & $\tau$ & Resampling & $K_x$ schedule & $K_\epsilon$ schedule \\
\midrule
1000D GMM & 1{,}000 & 2{,}000 & 10 & .95 & systematic & $100\!\to\!2;.05$ & $100\!\to\!2;.05$ \\
Chignolin & 2{,}000 & 600 & 2 & .95 & KL reshuffling & $200\!\to\!3;.01$ & $200\!\to\!3;.05$ \\
\bottomrule
\end{tabular}
\end{table}

\section{Proof of Flow Perturbation++}
\label{sec:fp_pp}

In this appendix we provide a complete derivation of the Flow Perturbation++ (FP++) estimator under exact inverse step maps, followed by the corresponding scope statement for numerically implemented reverse maps. Assume that every step map $\f_i$ is a $C^1$ diffeomorphism with a nonsingular Jacobian, that the normalized prior density satisfies $q(\bz)>0$ almost everywhere, and that the observable is integrable under the normalized target density, $\mathbb E_{\bx\sim p}|O(\bx)|<\infty$.
Consider a flow map $\f: \mathbb{R}^D \rightarrow \mathbb{R}^D$ obtained by composing $T$ numerical ODE integration steps:
\[
\f = \f_1 \circ \f_{2} \circ \cdots \circ \f_T .
\]
For an initial state $\bz \sim q(\bz)$ we define a trajectory
\[
\bx_1 = \bx, \quad \bx_i = \f_i(\bx_{i+1}), \quad \bx_{T+1} = \bz .
\]
Let $\hat{\epsilon}_1,\dots,\hat{\epsilon}_T$ be i.i.d. random vectors sampled uniformly from the unit sphere $S^{D-1}$. Define the FP++ action
\[
\Delta S(\bz,\hat{\epsilon}_1,\dots,\hat{\epsilon}_T)
=
- D \sum_{i=1}^T \log 
\left\|
\frac{\partial  \f_i^{-1}}{\partial \bx_i}
 \hat{\epsilon}_i
\right\|.
\]
Under these assumptions, we show that
\begin{equation}
\mathbb{E}_{\bx\sim p(\bx)}[O(\bx)]
=
\mathbb{E}_{\bz\sim q(\bz),\, \hat{\epsilon}_{1:T}}
\!\left[
O(\f(\bz))
\frac{p(\f(\bz))}{q(\bz)}
\exp\!\big(\Delta S(\bz,\hat{\epsilon}_{1:T})\big)
\right].
\label{eq:fp_pp_main}
\end{equation}
This identity holds for every test function $O$, hence it suffices to show that
\begin{equation}
\mathbb{E}_{\hat{\epsilon}_{1:T}}
\!\left[
\exp\!\big(\Delta S(\bz,\hat{\epsilon}_{1:T})\big)
\right]
=
\left|
\det
\frac{\partial  \f}{\partial \bz}
\right|.
\label{eq:fp_pp_core}
\end{equation}

\subsection{Reduction to a single-step identity}

Expanding the FP++ weight gives
\[
\exp(\Delta S)
=
\prod_{i=1}^T
\left\|
\frac{\partial  \f_i^{-1}}{\partial \bx_i}\hat{\epsilon}_i
\right\|^{-D}.
\]
Because the $\hat{\epsilon}_i$ are independent,
\[
\mathbb{E}_{\hat{\epsilon}_{1:T}}[\exp(\Delta S)]
=
\prod_{i=1}^T 
\mathbb{E}_{\hat{\epsilon}_i}
\left[
\left\|
\frac{\partial  \f_i^{-1}}{\partial \bx_i}\hat{\epsilon}_i
\right\|^{-D}
\right].
\]
Thus, the main statement reduces to proving the following single-step identity:

\begin{equation}
\mathbb{E}_{\hat{\epsilon}\sim S^{D-1}}
\left[\|A\hat{\epsilon}\|^{-D}\right]
=
|\det(A)|^{-1},
\qquad
A \in \mathbb{R}^{D\times D} \text{ invertible}.
\label{eq:single_step_identity}
\end{equation}

\subsection{Proof of the single-step identity}

Let $A = U\Sigma V^\top$ be the SVD of $A$, where $U$ and $V$ are orthogonal matrices and $\Sigma = \mathrm{diag}(\sigma_1,\dots,\sigma_D)$ contains the singular values of $A$. Orthogonal matrices preserve the uniform measure on the unit sphere, therefore $V^\top \hat{\epsilon}$ is uniformly distributed on $S^{D-1}$ as well. Using the rotational invariance of the Euclidean norm:
\[
\|A\hat{\epsilon}\|
=
\|U\Sigma V^\top \hat{\epsilon}\|
=
\|\Sigma v\|,
\qquad v = V^\top \hat{\epsilon}.
\]
Thus,
\[
\mathbb{E}_{\hat{\epsilon}}[\|A\hat{\epsilon}\|^{-D}]
=
\mathbb{E}_{v\sim S^{D-1}}
\left[
\left(
\sum_{j=1}^D \sigma_j^2 v_j^2
\right)^{-D/2}
\right].
\]

A classical result from integral geometry (obtainable via a change to ellipsoidal coordinates or by comparing Gaussian integrals in two ways) states that
\begin{equation}
\mathbb{E}_{v\sim S^{D-1}}
\left[
\left(
\sum_{j=1}^D \sigma_j^2 v_j^2
\right)^{-D/2}
\right]
=
\frac{1}{\prod_{j=1}^D \sigma_j }.
\label{eq:sphere_identity}
\end{equation}
Since $\prod_{j=1}^D \sigma_j = |\det(A)|$, equation \eqref{eq:single_step_identity} follows immediately.

\subsection{Completion of the FP++ proof}

Applying \eqref{eq:single_step_identity} to each integration step with 
$A_i = \frac{\partial  \f_i^{-1}}{\partial \bx_i}$ yields
\[
\mathbb{E}_{\hat{\epsilon}_{1:T}}[\exp(\Delta S)]
=
\prod_{i=1}^T
\left|
\det\!\left(
\frac{\partial  \f_i^{-1}}{\partial \bx_i}
\right)
\right|^{-1}
=
\prod_{i=1}^T
\left|
\det\!\left(
\frac{\partial  \f_i}{\partial \bx_{i+1}}
\right)
\right|.
\]
By the chain rule for Jacobians,
\[
\prod_{i=1}^T
\left|
\det\!\left(
\frac{\partial  \f_i}{\partial \bx_{i+1}}
\right)
\right|
=
\left|
\det\!\left(
\frac{\partial  \f}{\partial \bz}
\right)
\right|.
\]
Substituting into \eqref{eq:fp_pp_main} yields the FP++ importance sampling identity, completing the proof.
\qed

\phantomsection
\label{app:numerical_inverse_scope}
The proof assumes exact inverse-step directional derivatives. Finite differences introduce a numerical approximation, so exact unbiasedness does not directly extend to the implementation. Appendix~\ref{app:fd} assesses its agreement with the JVP reference for the implemented reverse map.

\section{Flow Perturbation++ variance reduction analysis}
\label{sec:fpplusplus_variance}
In this section, we analyze the log-domain variance of the Flow Perturbation++ (FP++) estimator relative to the full-flow Flow Perturbation (FP) estimator, conditional on a fixed flow trajectory. The variance comparison depends on the one-step marginal distributions and cross-step covariances.

Recall that in FP++, we use independent random perturbations for each flow step, whereas in FP, the same initial perturbation is propagated through all flow steps. For simplicity, denote the contribution of the $i$-th step as
\begin{equation}
X_i = - D \log \left\| \frac{\partial \f_i^{-1}}{\partial \bx_i} \hat{\epsilon}_i \right\|,
\end{equation}
where $\hat{\epsilon}_i$ is the independent unit random perturbation at step $i$ in FP++. In this appendix, write $J_i=\partial\f_i^{-1}/\partial\bx_i$ for the inverse-step Jacobian. Starting from $\hat{\epsilon}^{\mathrm{FP}}_0=\hat{\epsilon}\sim\operatorname{Unif}(S^{D-1})$, the direction in full-flow FP is propagated as
\begin{equation}
\hat{\epsilon}^{\mathrm{FP}}_i
=\frac{J_i\hat{\epsilon}^{\mathrm{FP}}_{i-1}}
{\|J_i\hat{\epsilon}^{\mathrm{FP}}_{i-1}\|},
\qquad
Y_i=-D\log\|J_i\hat{\epsilon}^{\mathrm{FP}}_{i-1}\|.
\label{eq:fp_direction_recursion}
\end{equation}
The total contributions are
\begin{equation}
\Delta S_{\mathrm{FP++}}=\sum_i X_i,\qquad
\Delta S_{\mathrm{FP}}=-D\log\|J_T\cdots J_1\hat{\epsilon}\|=\sum_i Y_i.
\label{eq:full_fp_exact_decomposition}
\end{equation}
Thus full-flow FP does not use an unchanged direction at every step.

\subsection{Single-step variance}

When each flow step acts on random unit vectors with identical distributions in FP and FP++, the single-step variances are
equal:
\begin{equation}
\operatorname{Var}[X_i] = \operatorname{Var}[Y_i].
\end{equation}

\subsection{Covariance structure}

The total variance for FP is
\begin{equation}
\operatorname{Var}\Big[\sum_i Y_i\Big]
=\sum_i\operatorname{Var}[Y_i]+2\sum_{i<j}\operatorname{Cov}[Y_i,Y_j].
\label{eq:fp_variance_exact}
\end{equation}
The covariance is
\begin{equation}
\operatorname{Cov}[Y_i, Y_j] = \mathbb{E}[Y_i Y_j] - \mathbb{E}[Y_i]\mathbb{E}[Y_j].
\end{equation}
Define
\begin{equation}
g_i(\hat{\epsilon}) = \log \|J_i \hat{\epsilon}\|, \quad
g_j(\hat{\epsilon}) = \log \|J_j \hat{\epsilon}\|.
\end{equation}
For a unit vector $\hat{\epsilon}$, we can write
\begin{equation}
g_i(\hat{\epsilon}) = \frac{1}{2} \log (\hat{\epsilon}^\top A_i \hat{\epsilon}), \quad
g_j(\hat{\epsilon}) = \frac{1}{2} \log (\hat{\epsilon}^\top A_j \hat{\epsilon}),
\end{equation}
where $A_i = J_i^\top J_i$ and $A_j = J_j^\top J_j$ are symmetric positive definite matrices.
Then, for the propagated directions in Eq.~\eqref{eq:fp_direction_recursion},
\begin{equation}
\operatorname{Cov}[Y_i,Y_j]
=D^2\operatorname{Cov}\big[g_i(\hat{\epsilon}^{\mathrm{FP}}_{i-1}),
g_j(\hat{\epsilon}^{\mathrm{FP}}_{j-1})\big].
\end{equation}
This covariance need not be non-negative for general step Jacobians.

\paragraph{Local covariance analysis.}
For nearby integration steps, aligned anisotropic stretching can give positive aggregate covariance. Consider small steps $J_i=I+h_iB_i+O(h^2)$, where $h_i=h\alpha_i$ with fixed $\alpha_i$, dimension, and number of steps. Let $S_i=(B_i+B_i^{\top})/2$ and $\widetilde S_i=S_i-\operatorname{tr}(S_i)I/D$.
\begin{lemma}[Local alignment condition]
\label{thm:strict_local_variance}
With uniformly bounded remainders,
\begin{equation}
\sum_{i<j}\operatorname{Cov}[Y_i,Y_j]
=\frac{2D}{D+2}\sum_{i<j}h_i h_j\langle\widetilde S_i,\widetilde S_j\rangle_F+O(h^3).
\label{eq:strict_log_variance_expansion}
\end{equation}
If $\kappa=\sum_{i<j}\alpha_i\alpha_j\langle\widetilde S_i,\widetilde S_j\rangle_F>0$, then $\sum_{i<j}\operatorname{Cov}[Y_i,Y_j]>0$ for all sufficiently small positive $h$. The sufficient condition applies in this small-step regime.
\end{lemma}
\begin{proof}
For a uniform initial direction $\hat{\epsilon}$, the propagated direction is $\hat{\epsilon}^{\mathrm{FP}}_{i-1}=\hat{\epsilon}+O(h)$, so $Y_i=-Dh_i\hat{\epsilon}^{\top}S_i\hat{\epsilon}+O(h^2)$. The spherical covariance identity is
\[
\operatorname{Cov}(\hat{\epsilon}^{\top}S_i\hat{\epsilon},\hat{\epsilon}^{\top}S_j\hat{\epsilon})
=\frac{2\langle\widetilde S_i,\widetilde S_j\rangle_F}{D(D+2)}.
\]
Multiplying by $D^2h_i h_j$ and summing gives Eq.~\eqref{eq:strict_log_variance_expansion}; the bounded remainder contributes $O(h^3)$. When $\kappa>0$, the sum is $c h^2+R(h)$ with $c=2D\kappa/(D+2)>0$ and $|R(h)|\le K h^3$ for some finite $K\ge0$. Choosing $h>0$ small enough that $Kh<c/2$ yields $\sum_{i<j}\operatorname{Cov}[Y_i,Y_j]\ge c h^2/2>0$.
\end{proof}

\subsection{Variance reduction in FP++}

For FP++, using independent perturbations, the covariance terms vanish:
under the matched one-step marginal condition in Section~C.1 and the positive aggregate covariance condition established by Lemma~\ref{thm:strict_local_variance},
\begin{equation}
\operatorname{Var}\Big[\sum_i X_i\Big]
=\sum_i\operatorname{Var}[X_i]
=\sum_i\operatorname{Var}[Y_i]
<\operatorname{Var}\Big[\sum_i Y_i\Big].
\label{eq:fppp_variance_exact}
\end{equation}
Thus, under these conditions, the total log-domain variance of the FP++ estimator is strictly smaller than that of FP. Intuitively, breaking the dependence between steps prevents positive aggregate correlation between contributions from inflating the overall variance. With matched one-step marginals, zero aggregate covariance gives equal variances; the ordering for general Jacobians depends on both the marginals and the covariances.

\subsection{Summary}

\begin{itemize}
\item The key difference is that FP++ uses independent random directions per step, whereas FP propagates its initial direction through the inverse flow.
\item The single-step variances are equal only when the marginal distributions match.
\item Removing cross-step dependence gives variance reduction under the stated marginal and covariance conditions; nearby-step alignment supplies a sufficient local condition.
\item Lower log-domain variance alone does not imply lower determinant or importance-weight variance. Appendix~\ref{app:direct_variance} reports direct determinant/weight diagnostics, while Appendix~\ref{appendix:variance_entropy} reports log-domain measurements.
\end{itemize}

\section{Gaussian mixture model benchmark \label{sec:GMM}}

In this work, we consider a Gaussian Mixture Model (GMM) to generate a multimodal test distribution. An \(n\)-dimensional GMM with \(k\) components represents the probability density as
\begin{equation}
p(\mathbf{x}) = \sum_{j=1}^{k} \pi_j \, \mathcal{N}(\mathbf{x} \mid \boldsymbol{\mu}_j, \Sigma_j) \,,
\end{equation}
where \(\pi_j\) is the mixture weight for the \(j\)-th Gaussian, and \(\mathcal{N}(\mathbf{x} \mid \boldsymbol{\mu}_j, \Sigma_j)\) denotes the corresponding multivariate normal distribution.

For our experiments, we adopt a two-component GMM (\(k=2\)) with the following setup:

- \textbf{Component Means (\(\boldsymbol{\mu}_j\))}: The first Gaussian is centered at \((-2, 0, \dots, 0)\), and the second at \((2, 0, \dots, 0)\), with the displacement applied only along the first dimension and all remaining dimensions set to zero.

- \textbf{Covariance Matrices (\(\Sigma_j\))}: The two components have distinct covariance structures:
  - Component 1 (\(j=1\)) uses a diagonal covariance matrix with positive entries drawn as \(0.01 + |\mathcal{N}(0,0.25)|\), ensuring positive definiteness.
  - Component 2 (\(j=2\)) employs a full covariance matrix: first, a diagonal matrix is sampled as above, then a random rotation (obtained via the QR decomposition of a random matrix) is applied to introduce correlations between dimensions.

- \textbf{Mixing Weights (\(\pi_j\))}: We set \(\pi_0 = 0.25\) and \(\pi_1 = 0.75\), creating an asymmetric mixture with the second component more heavily weighted.

The density can equivalently be expressed as a Boltzmann distribution,
\begin{equation}
p(\mathbf{x}) = \exp(-E(\mathbf{x})),
\end{equation}
where \(E(\mathbf{x})\) denotes the energy of configuration \(\mathbf{x}\).  

This particular GMM presents a challenging scenario for sampling, with two well-separated modes and heterogeneous covariance structures, making it a suitable benchmark for testing and validating advanced Boltzmann sampling techniques.

\section{Brute-force and stochastic estimation of CNF Jacobian traces}\label{appendix:CNFJacobian}

Continuous Normalizing Flows (CNFs) generalize standard Normalizing Flows by defining the mapping 
\(\f: \bz \to \bx\) through an ordinary differential equation (ODE):
\begin{equation}
\frac{d\bx_t}{dt} = \vvec(\bx_t, t), \quad \bx_0 = \bz,
\end{equation}
where \(\vvec(\bx,t)\) denotes the velocity field governing the transformation.  
The log-determinant of the flow Jacobian can then be expressed as an integral over the divergence of \(\vvec\):
\begin{equation}
\log \Bigl|\det \frac{\partial \f}{\partial \bz} \Bigr| = \int_0^1 \nabla \cdot \vvec(\bx_t, t)\, dt,
\end{equation}
with the divergence given by the trace of the Jacobian \(\partial \vvec / \partial \bx\).

We outline two standard approaches for computing this Jacobian trace: a direct (brute-force) method and a stochastic method using the Hutchinson estimator.

\subsection{Direct Jacobian construction via automatic differentiation}

Let \(\vvec: \mathbb{R}^D \to \mathbb{R}^D\) be differentiable.  
A full Jacobian \(\partial \vvec / \partial \bx\) can be obtained using automatic differentiation frameworks (e.g., PyTorch):

\begin{enumerate}
\item For each output component \(v_i(\bx)\), define the unit vector \(\mathbf{e}_i \in \mathbb{R}^D\) with a 1 in the \(i\)-th position and zeros elsewhere, such that
\[
v_i(\bx) = \mathbf{e}_i^\top \vvec(\bx).
\]

\item Compute the gradient \(\nabla_\bx v_i(\bx)\) via backpropagation.  
This yields the \(i\)-th row of the Jacobian:
\[
\nabla_\bx v_i(\bx) = \Bigl[ \frac{\partial v_i}{\partial x_1}, \frac{\partial v_i}{\partial x_2}, \dots, \frac{\partial v_i}{\partial x_D} \Bigr].
\]

\item Stack all rows to assemble the full Jacobian:
\[
\frac{\partial \vvec}{\partial \bx} =
\begin{bmatrix}
\nabla_\bx v_1(\bx) \\ \nabla_\bx v_2(\bx) \\ \vdots \\ \nabla_\bx v_D(\bx)
\end{bmatrix}.
\]
\end{enumerate}

In practice, PyTorch provides \texttt{jacrev} to efficiently compute this Jacobian.

\subsection{Hutchinson estimator for trace approximation}

The Hutchinson estimator offers a stochastic alternative for estimating \(\mathrm{Tr}(\partial \vvec / \partial \bx)\) without explicitly forming the full Jacobian:

\begin{enumerate}
\item Sample a random vector \(\uvec \in \mathbb{R}^D\) from a standard normal or Rademacher distribution.

\item Compute the gradient of the scalar \(\uvec^\top \vvec(\bx)\) w.r.t.\ \(\bx\):
\[
\mathbf{g} = \nabla_\bx (\uvec^\top \vvec(\bx)) = \frac{\partial (\uvec^\top \vvec)}{\partial \bx}.
\]

\item Form the quadratic form \(\uvec^\top \mathbf{g} = \uvec^\top (\partial \vvec / \partial \bx)\, \uvec\), which is an unbiased estimator of the trace.

\item Repeat for \(N\) independent random vectors \(\uvec_i\) and average:
\[
\mathrm{Tr}\Bigl(\frac{\partial \vvec}{\partial \bx}\Bigr) \approx \frac{1}{N} \sum_{i=1}^N \uvec_i^\top \frac{\partial \vvec}{\partial \bx} \, \uvec_i.
\]
\end{enumerate}

Each sample requires one backpropagation pass. Averaging over multiple \(\uvec_i\) reduces estimator variance, providing an efficient method for high-dimensional CNFs.

\section{Computational cost of JVP/VJP-based estimators}
\label{app:jvp_cost}

\subsection{FP and FP++ implementations: FD, JVP, and VJP}
\label{app:fp_jvp_vjp}

The inverse--Jacobian--vector product appearing in Eq.~\ref{eq:fp_++}, 
$J_k^{-1} \epsilon_k = (\partial \mathbf{f}_k^{-1}/\partial \mathbf{x}_k)\, \epsilon_k$, 
can be implemented in three ways. In practice, the reverse integration approximates $\mathbf f_k^{-1}$; a JVP of this implemented map is exact for that map, not necessarily for the true inverse (Appendix~\ref{app:numerical_inverse_scope}).

\paragraph{Finite-difference (FD) approximation.}  
The product is approximated by perturbing the inverse flow:
\[
J_k^{-1} \epsilon_k \approx \frac{\mathbf{f}_k^{-1}(\mathbf{x}_k + \delta \epsilon_k) - \mathbf{f}_k^{-1}(\mathbf{x}_k - \delta \epsilon_k)}{2 \delta}.
\]

\paragraph{Forward-mode Jacobian-vector product (JVP).}  
The JVP computes the exact directional derivative along $\epsilon_k$:
\[
\mathrm{JVP}(\mathbf{f}_k^{-1}, \mathbf{x}_k, \epsilon_k) 
= \left. \frac{d}{d\eta} \mathbf{f}_k^{-1}(\mathbf{x}_k + \eta \epsilon_k) \right|_{\eta=0} 
= J_k^{-1} \epsilon_k.
\]

\paragraph{Reverse-mode vector-Jacobian product (VJP).}  
Given a cotangent $\epsilon_k$ at the output, the VJP computes
\[
\mathrm{VJP}(\mathbf{f}_k^{-1}, \mathbf{x}_k, \epsilon_k) = \epsilon_k^\top J_k^{-1}.
\]
For the same probe, the VJP and JVP norms need not be equal. Their squared norms involve the inverse Jacobian and its transpose in opposite orders; these matrices have the same eigenvalues. Uniform spherical probes therefore give identical norm distributions, rather than pointwise equality.

\paragraph{Delta entropy computation.}  
For all three implementations, the contribution to the flow-based entropy change at step $k$ is
\[
\Delta S_k = -\frac{D}{2} \log \frac{\|J_k^{-1} \epsilon_k\|^2}{\|\epsilon_k\|^2},
\]

\subsection{FD-based implementations in the main text}
\label{app:fd_main}

In the main text, FP and FP++ are implemented using \emph{finite-difference} (FD) estimators. 
Here, Jacobian information is approximated by evaluating the inverse flow at perturbed inputs. This requires:
\begin{itemize}
    \item 1 ODE pass for trajectory generation, and
    \item 2 additional ODE passes for entropy estimation,
\end{itemize}
yielding a total of 3 ODE passes per SMC step. 

\subsection{JVP/VJP-based implementations}
\label{app:jvp_vjp_main}

In this appendix and in Table~\ref{tab:jvp_cost}, we report runtimes for implementations that use explicit vector--Jacobian (VJP) or Jacobian--vector (JVP) products.  
In this regime, entropy-related terms are computed via automatic differentiation without perturbing the inverse flow. As a result, FP and FP++ require:
\begin{itemize}
    \item 1 ODE pass for trajectory propagation, and
    \item 1 ODE pass for the associated JVP/VJP computation,
\end{itemize}
yielding a total of 2 ODE passes per SMC step. In the reported comparison, the G2/R2 and G10/R10 Hutchinson variants use two and ten stochastic probes, respectively, resulting in more ODE passes and longer wall-clock times.

\subsection{Wall-clock times and interpretation}
\label{app:wall_clock}

Table~\ref{tab:jvp_cost} summarizes both the theoretical (ODE counts) and empirical (wall-clock) computational cost for different Jacobian estimators under the JVP/VJP regime.  
The columns \textbf{VJP Time} and \textbf{JVP Time} report measured wall-clock times for explicit vector--Jacobian and Jacobian--vector implementations, respectively.  
The results indicate that, for JVP/VJP-based implementations, the dominant computational cost arises from the number of ODE solver evaluations.

\begin{table}[h]
    \centering
    \caption{Computational cost of different Jacobian estimators.
    Cost (\# ODE passes) counts the number of ODE solver evaluations per SMC step, including trajectory propagation and entropy computation.
    VJP Time and JVP Time report the total wall-clock time incurred by implementations based on explicit vector--Jacobian or Jacobian--vector products, respectively.}
    \label{tab:jvp_cost}
    \small
    \begin{tabular}{lccc}
        \toprule
        Method & Cost (\# ODE passes) & VJP Time & JVP Time \\
        \midrule
        Hutchinson G2 / R2   & 3  & 30.4\,h  & 48.9\,h \\
        Hutchinson G10 / R10 & 11 & 100\,h   & 176\,h  \\
        FP (single-step)    & 2  & 23.8\,h  & 30.8\,h \\
        FP++ (multi-step)   & 2  & 23.8\,h  & 30.8\,h \\
        \bottomrule
    \end{tabular}
\end{table}

\subsection{Step count, intermediate states, and memory}
\label{app:state_memory}

For multiple probes at a fixed input, one primal linearization can be constructed and its linear map reused for all probe directions. This requires one primal evaluation and one linear-map application per probe, rather than two perturbed inverse evaluations per finite-difference probe. Retaining the reusable linearization introduces a memory trade-off. The timings above cover the evaluated implementations; the reuse strategy requires a separate runtime measurement.

The ODE-equivalent counts in Tables~\ref{table:compare} and~\ref{tab:jvp_cost} treat one complete trajectory solve as one pass. They therefore hide solver-dependent constants: wall-clock cost grows with the number of integration steps and stages per step, and adaptive solvers add control-flow and rejected-step overhead. State management also matters. A direct per-step implementation may retain $T+1$ trajectory states for each particle, requiring $O(MTD)$ state storage for $M$ particles in $D$ dimensions. Streaming reverse evaluation or checkpointing can reduce resident state storage toward $O(MD)$, or an intermediate checkpoint-dependent cost, by recomputing portions of the trajectory. JVP/VJP implementations may additionally retain or reconstruct an automatic-differentiation graph, whereas finite differences use forward evaluations and avoid that graph at the cost of two perturbed solves per probe. Table~\ref{tab:ad2_cost} provides a runtime and memory comparison for alanine dipeptide under the evaluated solver settings. A systematic comparison across solver families, tolerances, and step counts remains outside the scope of these measurements.

\section{Molecular dynamics simulations of Chignolin mutant \label{appendix:MD}}

We performed molecular dynamics (MD) simulations to investigate the behavior of the Chignolin Mutant. A total of six independent trajectories were generated, each consisting of \(10^7\) integration steps with a time step of 1 fs. All simulations were carried out at a temperature of 300 K, employing the CHARMM22 force field together with the implicit OBC2 solvent model. These conditions are consistent with those used for producing the training dataset. 

\begin{figure}[h]
    \centering
    \includegraphics[width=0.5\linewidth]{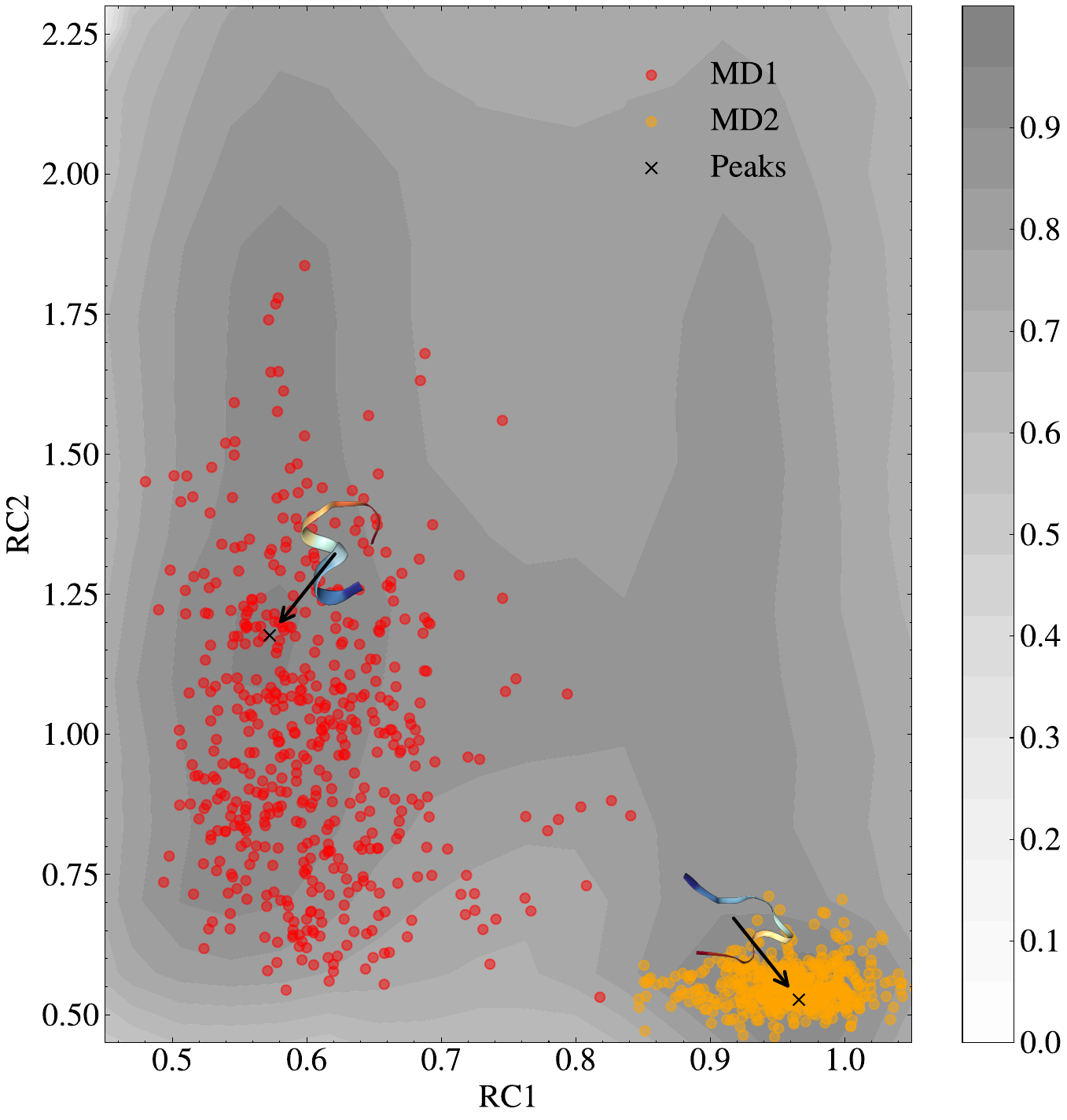}
    \caption{Representative MD trajectories of Chignolin Mutant visualized on the reaction coordinate plane.}
    \label{fig:MD_chigolin}
\end{figure}

Across all runs, the system remained confined within one of its metastable conformational states. Figure~\ref{fig:MD_chigolin} illustrates two representative trajectories projected onto the chosen reaction coordinate (RC) plane.

\section{Additional experimental results}
\label{appendix:extra_results}

This section provides supplementary experimental results that further evaluate the accuracy, stability, and convergence behavior of the proposed estimators.

\subsection{Variance of the log-domain estimator}
\label{appendix:variance_entropy}
We evaluate the variance of the logged Jacobian-correction estimates across methods and dimensionalities using 1000 samples. These quantities provide the stable numerical representation used in computation, whereas determinant-level unbiasedness pertains to their exponentiated positive factors. Table~\ref{tab:variance_entropy} summarizes the results.
\begin{table}[ht]
\centering
\caption{Variance of logged Jacobian-correction estimates (1000 samples) across methods and dimensionalities.}
\label{tab:variance_entropy}
\small
\begin{tabular}{lccc}
\toprule
Method & GMM 10D & GMM 100D & GMM 1000D \\
\midrule
BF Jacobian & 0.9265 & 1.5931 & 11.0544 \\
\midrule
FP1 & 1.6886 & 10.2365 & 37.6025 \\
FP2 & 1.4847 & 8.0238 & 35.3451 \\
FP5 & 1.1487 & 6.5414 & 31.2574 \\
FP10 & 0.9597 & 5.3718 & 29.4457 \\
\midrule
FP1++ & 0.9456 & 1.8040 & 11.4718 \\
FP2++ & 0.9393 & 1.6681 & 11.3622 \\
FP5++ & 0.9259 & 1.6174 & 11.1046 \\
FP10++ & 0.9255 & 1.6193 & 11.1099 \\
\midrule
G1 & 1.1992 & 2.7067 & 12.8340 \\
G2 & 1.1018 & 2.1713 & 12.0969 \\
G5 & 0.9930 & 1.8662 & 11.5580 \\
G10 & 0.9503 & 1.7211 & 11.2355 \\
\midrule
R1 & 0.9394 & 1.7396 & 11.3268 \\
R2 & 0.9378 & 1.6757 & 11.1838 \\
R5 & 0.9266 & 1.6275 & 11.1346 \\
R10 & 0.9281 & 1.6109 & 11.0810 \\
\bottomrule
\end{tabular}
\end{table}
\paragraph{Setup.}
We consider Gaussian mixture models (GMMs) with dimensions ranging from 10D to 1000D. BF Jacobian denotes the exact computation and serves as ground truth. 
Methods with prefix FP use finite-difference estimators (FP/FP++), where the number indicates the number of probes per step (e.g., FP1 uses one probe). 
Methods denoted as G$k$ and R$k$ correspond to Hutchinson estimators using Gaussian and Rademacher probes, respectively.

\paragraph{Results.}
We observe that stochastic estimators such as Hutchinson exhibit increasing variance as dimensionality grows, especially under small probe budgets. Increasing the number of probes generally reduces variance but incurs higher computational cost.

In contrast, FP++ consistently achieves log-domain variance close to the BF Jacobian across all dimensions, even with a single probe ($N=1$). Notably, FP++ maintains stable log-domain variance as dimensionality increases, whereas the single-step FP estimator exhibits significantly higher variance, particularly in high dimensions.

These results demonstrate that the multi-step aggregation in FP++ achieves significantly lower log-domain variance than FP and low-probe Hutchinson-based estimators in these experiments, while reaching a variance level comparable to the brute-force Jacobian, making it a more accurate and robust estimator under the same computational budget. Appendix~\ref{app:direct_variance} separately evaluates determinant and weight variance.

\subsection{Convergence analysis of FP++}
\label{app:convergence}

We study the convergence behavior of FP++ with respect to the number of independent runs $M$. 
The estimation error is defined as the average absolute difference between the estimated Jacobian determinant and the ground-truth value, averaged over both independent runs and a fixed set of samples (number = 5).

\begin{figure}[ht]
\centering
\begin{minipage}{0.48\textwidth}
    \centering
    \includegraphics[width=\linewidth]{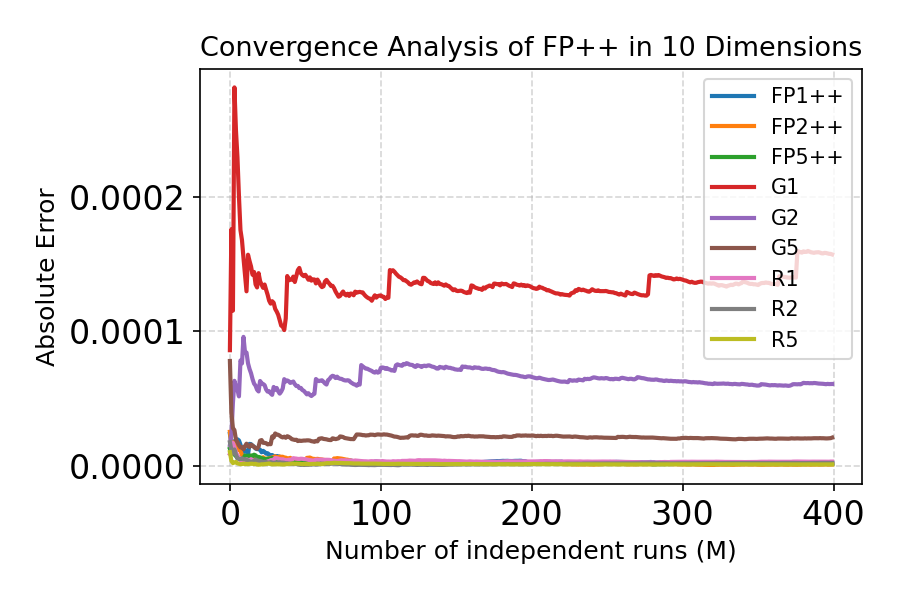}
    \caption*{(a) 10D}
\end{minipage}
\hfill
\begin{minipage}{0.48\textwidth}
    \centering
    \includegraphics[width=\linewidth]{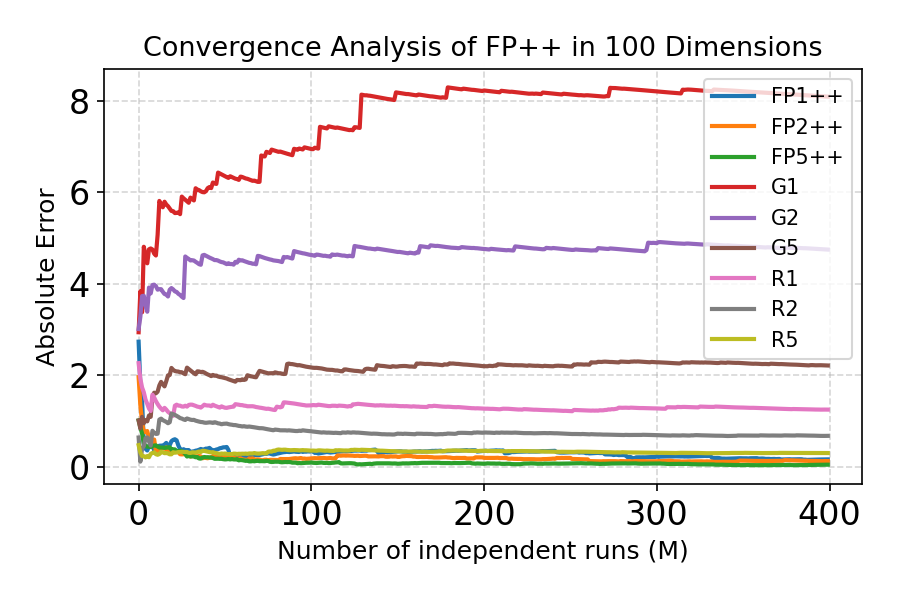}
    \caption*{(b) 100D}
\end{minipage}

\begin{minipage}{0.55\textwidth}
    \centering
    \includegraphics[width=\linewidth]{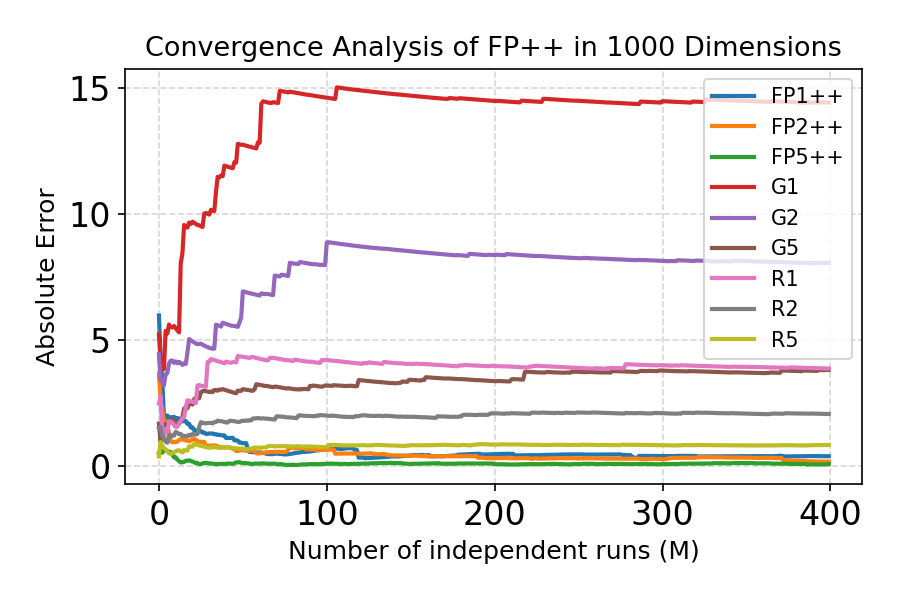}
    \caption*{(c) 1000D}
\end{minipage}
\caption{Running-average error as the number of independent repetitions $M$ increases. Curves stabilize at method-dependent residual errors in 10D, 100D, and 1000D.}
\label{fig:fp_convergence}
\end{figure}

\paragraph{Results.}
Figure~\ref{fig:fp_convergence} reports the running-average errors in 10D, 100D, and 1000D Gaussian mixture models. As $M$ grows, the curves settle toward method-dependent plateaus rather than decreasing monotonically in every case. These plateaus separate finite-repetition noise from residual estimator or numerical error relative to the brute-force reference and complement the log-domain variance comparison in Section~\ref{appendix:variance_entropy}.

\FloatBarrier
\subsection{Effect of finite-difference step size}
\label{app:fd}

We analyze the sensitivity of FP++ to the finite-difference step size $\delta$ and evaluate its numerical stability across different dimensionalities. We additionally compare finite-difference FP++ with a Jacobian-vector-product (JVP) implementation. This comparison isolates finite-difference error relative to the implemented reverse map, while Appendix~\ref{app:numerical_inverse_scope} characterizes the complementary solver-inversion error.

\paragraph{Setup.}
We evaluate FP++ with different choices of $\delta$, where the notation FP++($\delta$) denotes the estimator using step size $\delta$. Experiments are conducted on Gaussian mixture models (10D, 100D, 1000D) and a molecular system (Chignolin mutant).

\begin{figure}[htbp]
\centering

% ===== Row 1 =====
\begin{minipage}{0.48\textwidth}
    \centering
    \includegraphics[width=\linewidth]{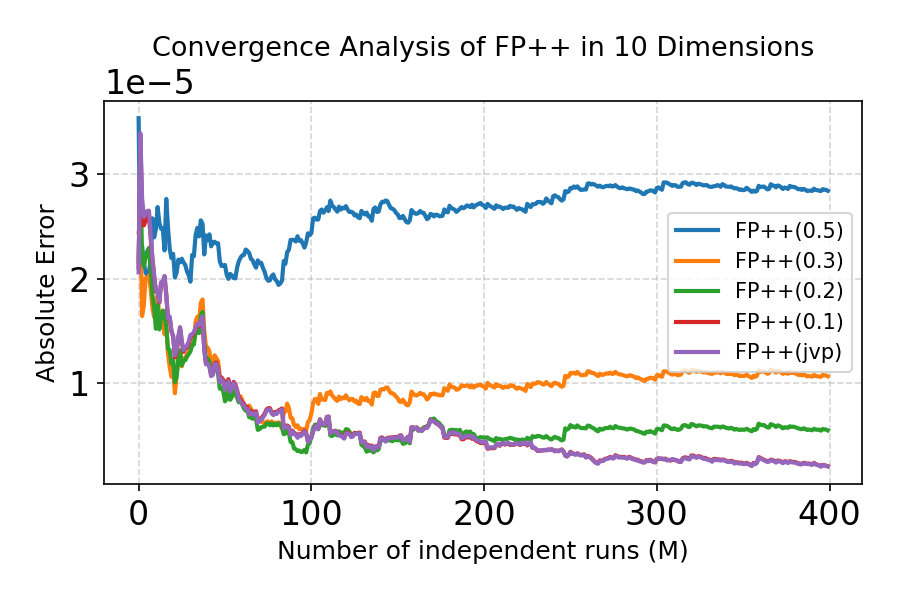}
    \caption*{(a) 10D GMM}
\end{minipage}
\hfill
\begin{minipage}{0.48\textwidth}
    \centering
    \includegraphics[width=\linewidth]{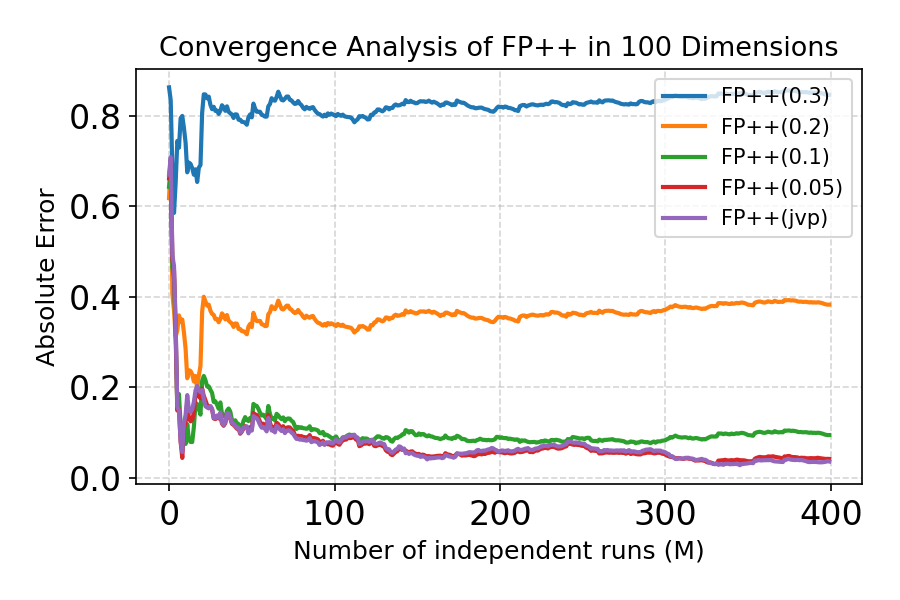}
    \caption*{(b) 100D GMM}
\end{minipage}

% ===== Row 2 =====
\begin{minipage}{0.48\textwidth}
    \centering
    \includegraphics[width=\linewidth]{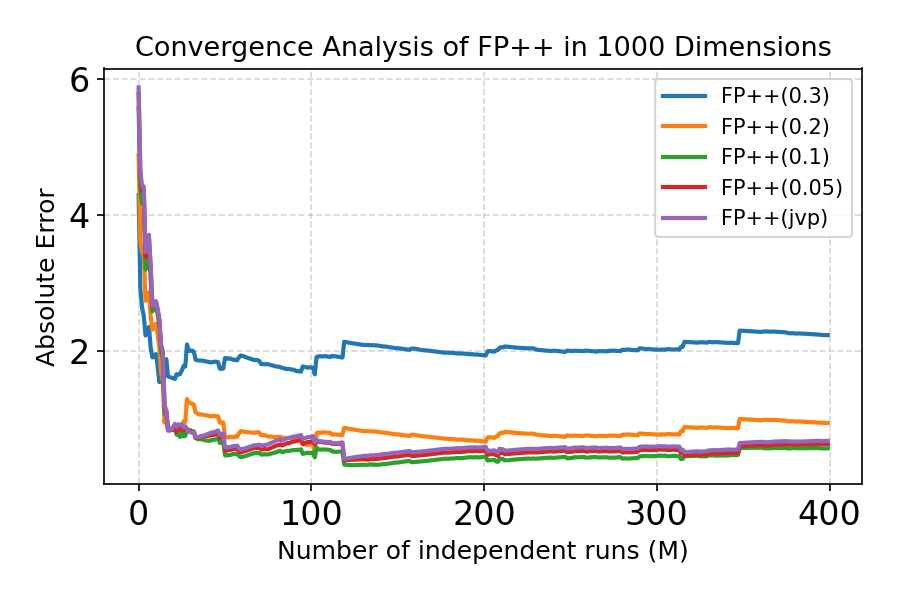}
    \caption*{(c) 1000D GMM}
\end{minipage}
\hfill
\begin{minipage}{0.48\textwidth}
    \centering
    \includegraphics[width=\linewidth]{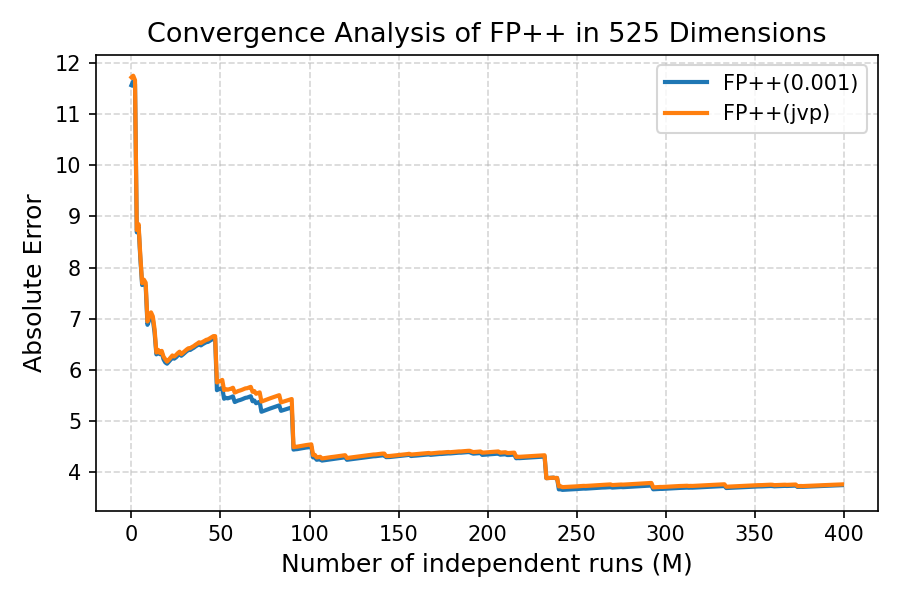}
    \caption*{(d) Chignolin mutant}
\end{minipage}

\caption{Effect of finite-difference step size $\delta$ on FP++. As $\delta$ decreases, finite-difference FP++ converges to the JVP-based result across different dimensionalities and systems.}
\label{fig:fp_delta}
\end{figure}

\paragraph{Results.}
The results are shown in Figure~\ref{fig:fp_delta}. We observe the following:

\begin{itemize}
    \item In low-dimensional settings (10D), finite-difference FP++ closely matches the JVP result even with relatively large step sizes (e.g., $\delta = 0.1$).
    
    \item In higher dimensions (100D and 1000D), smaller step sizes (e.g., $\delta = 0.05$) are sufficient to achieve stable and consistent estimates.
    
    \item In the more complex Chignolin system, a smaller step size ($\delta = 0.001$) is required to maintain high numerical accuracy and agreement with JVP.
\end{itemize}

Overall, as $\delta$ decreases, the finite-difference FP++ estimate approaches the JVP-based result before entering a roundoff-sensitive regime.

In all main experiments, we adopt $\delta = 0.001$, which lies in the visually stable regime of Figure~\ref{fig:fp_delta} and agrees closely with the JVP reference at the plotted resolution.

\subsection{Stability on Chignolin system}
\label{appendix:Chignolin}
We evaluate the stability of FP++ on the Chignolin system across 10 independent runs. 
We report two metrics:

\begin{itemize}
    \item \textbf{Proportion}: the fraction of particles located in the right basin, corresponding to the $\beta$-hairpin configuration;
    \item \textbf{\# Preserved}: the number of preserved lineages after resampling, measuring particle diversity.
\end{itemize}

\paragraph{Results.}
The results are summarized in Table~\ref{tab:chignolin_fpplusplus_seeds}. We observe that FP++ exhibits highly consistent performance across runs.

Specifically, the proportion of particles in the target basin remains stable (mean $0.3658$ with low standard deviation $0.0146$), indicating reliable sampling of the metastable state. 
At the same time, the number of preserved lineages remains consistently high (mean $1473.8$), suggesting that FP++ maintains strong particle diversity after resampling.

These results demonstrate that FP++ achieves both stable sampling behavior and effective diversity preservation in a realistic molecular system.

\begin{table}[ht]
\centering
\caption{Stability evaluation on the Chignolin system across 10 independent runs.}
\label{tab:chignolin_fpplusplus_seeds}
\small
\begin{tabular}{l l c c c c c c}
\toprule
Method & Metric & 1 / 6 & 2 / 7 & 3 / 8 & 4 / 9 & 5 / 10 & Mean $\pm$ Std \\
\midrule
% 第一组数据：Proportion
{FP++} & {Proportion} & 0.3640 & 0.3745 & 0.3600 & 0.3455 & 0.3540 & {0.3658 $\pm$ 0.0146} \\
& & 0.3855 & 0.3775 & 0.3580 & 0.3515 & 0.3875 & \\
\midrule
% 第二组数据：# Preserved
{FP++} &{\# Preserved} & 1463 & 1495 & 1461 & 1451 & 1498 & {1473.8 $\pm$ 18.0} \\
& & 1477 & 1491 & 1447 & 1480 & 1475 & \\
\bottomrule
\end{tabular}
\end{table}

Matched ten-run statistics for all estimators are reported in Table~\ref{tab:chignolin_stability_compact}. FP++ has the mean closest to the target $0.385$ and the smallest basin-fraction standard deviation among the tested estimators. FP occasionally approaches the target but varies more strongly across runs, while all Hutchinson variants underestimate the target on average.

\begin{table}[ht]
\centering
\caption{Chignolin repeated-run statistics under matched settings. All methods use ten independent runs.}
\label{tab:chignolin_stability_compact}
\small
\begin{tabular}{@{}lccc@{}}
\toprule
Method & Basin fraction & Range & \# Preserved \\
\midrule
FP++ & $0.3658\pm0.0146$ & $0.3455$--$0.3875$ & $1473.8\pm18.0$ \\
FP & $0.4450\pm0.0510$ & $0.3320$--$0.4830$ & $1432.7\pm28.3$ \\
R2 & $0.2330\pm0.0168$ & $0.2115$--$0.2675$ & $1529.3\pm25.4$ \\
G2 & $0.2136\pm0.0685$ & $0.1755$--$0.4065$ & $1315.7\pm39.7$ \\
R10 & $0.2457\pm0.0369$ & $0.2100$--$0.3210$ & $1541.1\pm29.8$ \\
G10 & $0.2303\pm0.0394$ & $0.1975$--$0.3260$ & $1458.4\pm19.0$ \\
\bottomrule
\end{tabular}
\end{table}

Table~\ref{tab:chignolin_metrics} reports complementary Chignolin diagnostics from representative single runs. Importance ESS is measured before resampling; lineage ESS accounts for common ancestry after resampling. Unique/$N$ is the fraction of distinct surviving lineages. Energy-W2 is measured in $k_BT$ and torsion-W2 in radians.
\begin{table}[ht]
\centering
\caption{Representative-run Chignolin diagnostics with 2,000 particles.}
\label{tab:chignolin_metrics}
\small
\begin{tabular}{@{}lcccccc@{}}
\toprule
Method & Basin & Energy-W2 & Torsion-W2 & Lin. ESS/N & Imp. ESS/N & Unique/N \\
\midrule
FP & 0.4750 & 1.6512 & 3.0853 & 0.036647 & 0.001190 & 0.7140 \\
FP++ & 0.3640 & 2.7330 & 2.7349 & 0.043632 & 0.003499 & 0.7315 \\
R2 & 0.2255 & 30.1826 & 2.7397 & 0.084090 & 0.002502 & 0.7685 \\
R10 & 0.2240 & 6.7210 & 2.6891 & 0.062073 & 0.001761 & 0.7675 \\
G2 & 0.1945 & 31.0080 & 2.8770 & 0.014943 & 0.000534 & 0.6650 \\
G10 & 0.2220 & 11.1386 & 2.7375 & 0.036490 & 0.002464 & 0.7225 \\
\bottomrule
\end{tabular}
\end{table}

\subsection{Fixed-input determinant and weight variance}
\label{app:direct_variance}
To separate estimator randomness from configuration variation, we use a two-well GMM with an analytically available determinant: $\tfrac12\mathcal N(-1.5\mathbf d,0.9^2\mathbf I)+\tfrac12\mathcal N(1.5\mathbf d,0.9^2\mathbf I)$, where $\mathbf d=(1,\ldots,1)/\sqrt D$. An exact quantile transport is decomposed into eight invertible steps; inversion errors are below $6.7\times10^{-15}$. We evaluate 16 fixed latent quantiles per dimension and repeat each estimator 5,000 times per input. Here $\widehat J/J$ is also the induced importance-weight ratio. Table~\ref{tab:direct_variance} summarizes the diagnostics.
\begin{table}[ht]
\centering
\caption{Fixed-input determinant and importance-weight diagnostics, aggregated over 16 fixed inputs per dimension and 5,000 estimator repetitions per input.}
\label{tab:direct_variance}
\small
\begin{tabular}{c l c c c c}
\toprule
$D$ & Method & Mean & Var. & CV$^2$ & ESS/$N$ \\
\midrule
10 & FP1 & 0.98836 & 15.20617 & 13.72306 & 0.44239 \\
10 & FP2 & 0.98621 & 7.39192 & 7.32949 & 0.53670 \\
10 & FP10 & 1.01540 & 5.91418 & 5.11221 & 0.67136 \\
10 & FP1++ & 0.99862 & 0.15049 & 0.15189 & 0.88678 \\
10 & FP2++ & 0.99951 & 0.07103 & 0.07135 & 0.93870 \\
10 & FP10++ & 1.00008 & 0.01364 & 0.01364 & 0.98678 \\
\addlinespace
100 & FP1 & 0.96615 & 90.14904 & 42.21219 & 0.34038 \\
100 & FP2 & 0.96248 & 35.38961 & 25.97286 & 0.42131 \\
100 & FP10 & 0.98688 & 38.01803 & 31.88503 & 0.52187 \\
100 & FP1++ & 1.00348 & 0.24940 & 0.24687 & 0.84224 \\
100 & FP2++ & 1.00150 & 0.12210 & 0.12154 & 0.90734 \\
100 & FP10++ & 1.00013 & 0.02134 & 0.02131 & 0.97979 \\
\addlinespace
1000 & FP1 & 0.95217 & 19.12200 & 21.79667 & 0.35077 \\
1000 & FP2 & 0.94331 & 6.79784 & 9.22970 & 0.44994 \\
1000 & FP10 & 0.94874 & 2.52255 & 3.48656 & 0.56135 \\
1000 & FP1++ & 0.99910 & 0.33129 & 0.33086 & 0.82973 \\
1000 & FP2++ & 1.00030 & 0.17903 & 0.17742 & 0.89120 \\
1000 & FP10++ & 1.00077 & 0.02629 & 0.02609 & 0.97576 \\
\bottomrule
\end{tabular}
\end{table}
FP1++ has lower determinant variance than FP1 for all 16 fixed inputs in each dimension. The median FP1/FP1++ variance ratios are 19.99$\times$, 19.03$\times$, and 22.82$\times$ in 10D, 100D, and 1000D, respectively. These fixed-input measurements show that FP++ reduces determinant and importance-weight variance relative to FP under the evaluated benchmark conditions.

\subsection{Feynman--Kac correctors on GMM benchmark}

We evaluate Feynman--Kac corrector (FKC)-based methods\cite{skreta2025feynmankac} on a 10D Gaussian Mixture Model (GMM) with an intentionally imbalanced target distribution. 
The metric \textit{neg ratio} denotes the fraction of samples whose first dimension is less than zero, corresponding to one mode of the mixture. The ground-truth target ratio is 0.25, reflecting the desired 25/75 mixture proportion.
\begin{figure}[htbp]
\centering

\includegraphics[width=0.7\linewidth]{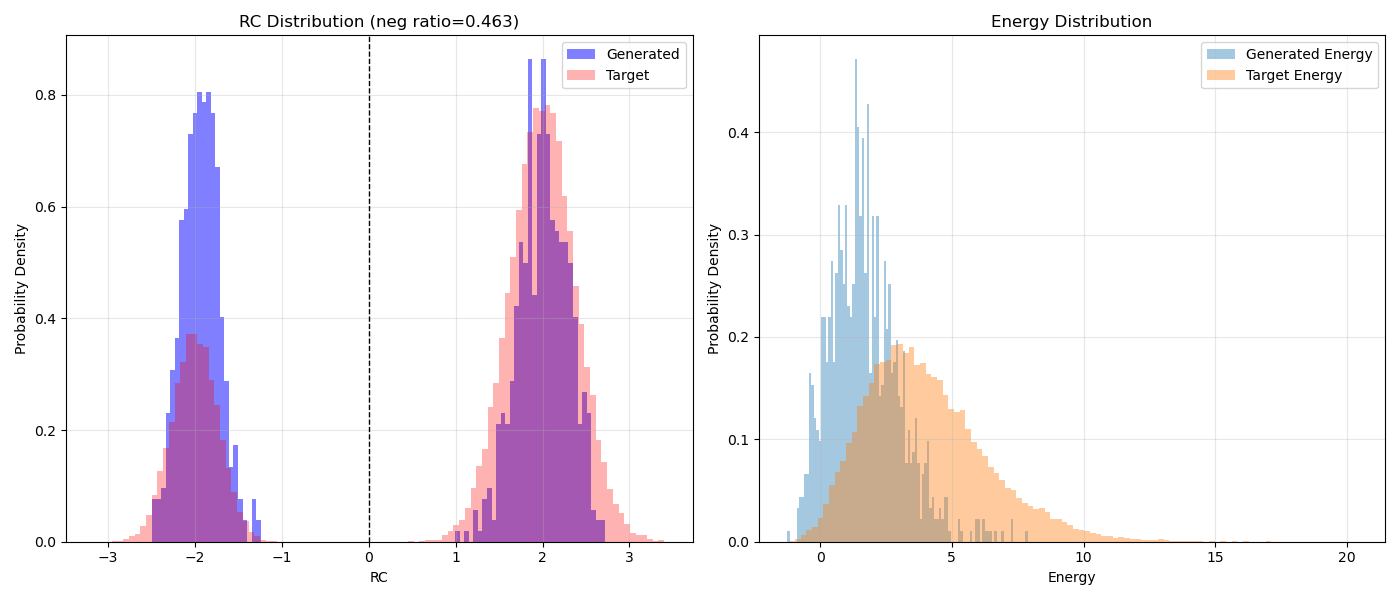}
\includegraphics[width=0.7\linewidth]{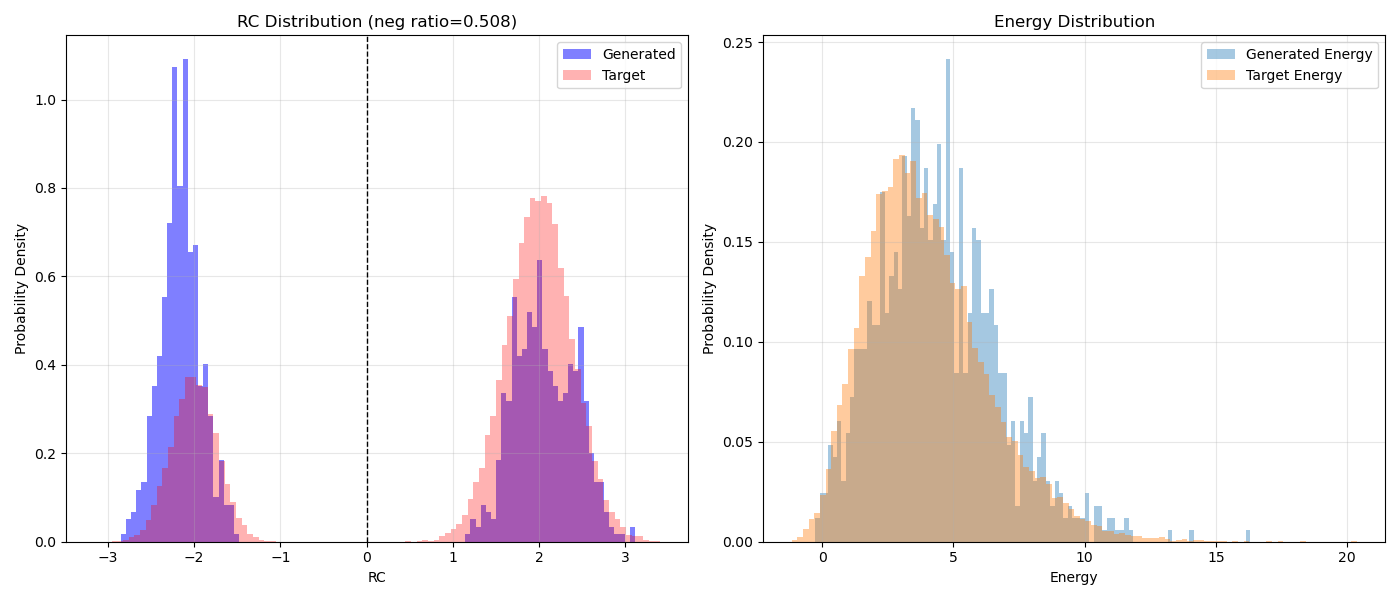}

\includegraphics[width=0.7\linewidth]{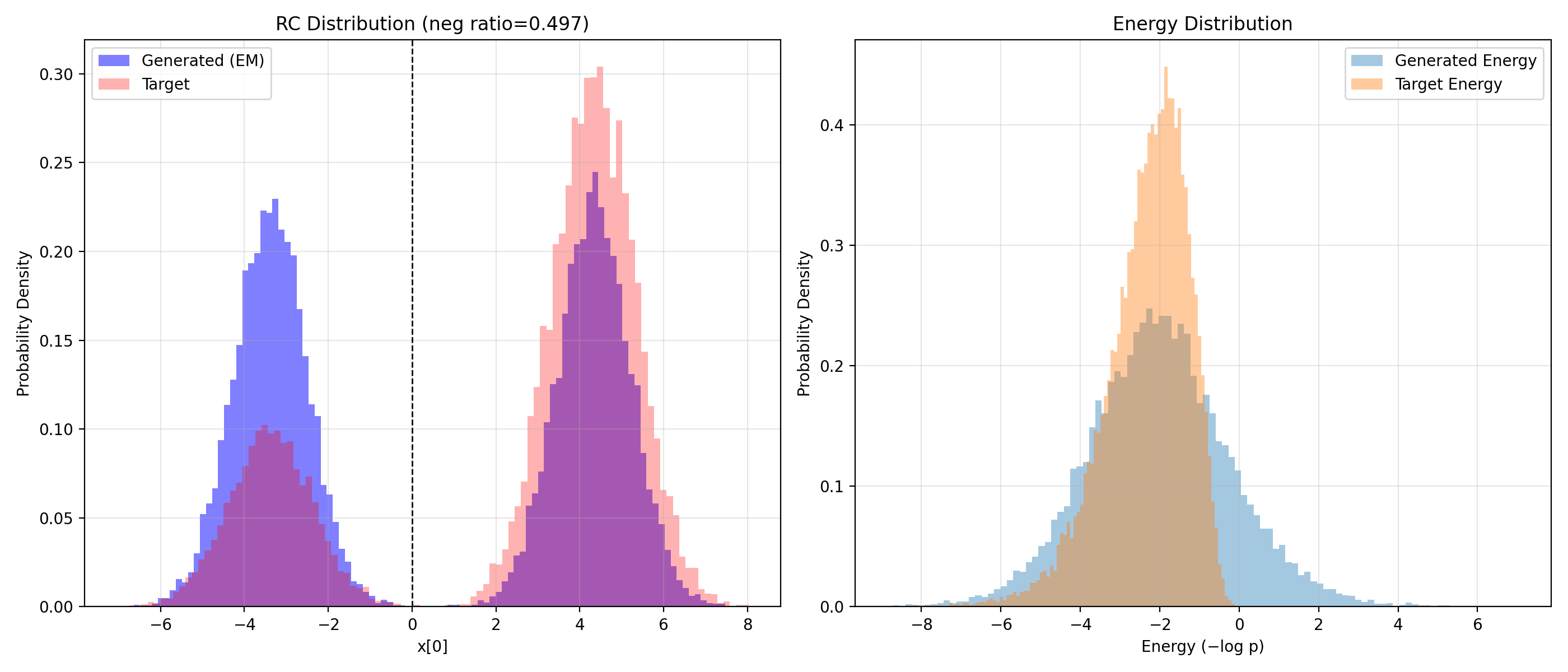}

\caption{FKC and RNE results on the 10D GMM benchmark. The target neg ratio is 0.25 (true mixture proportion). FKC and RNE methods fail to correct the inherited bias of the proposal distribution, while FP++ successfully recovers the correct mixture proportions.}
\label{fig:fkc_results}
\end{figure}
\paragraph{Setup.}
We consider two settings: (i) annealed SDE sampling with FKC correction, and (ii) reward-tilted sampling with FKC. Additionally, we compare against RNE-based methods. All methods are evaluated under the same proposal distribution.

\paragraph{Results.}
Figure~\ref{fig:fkc_results} summarizes the results. We observe that FKC-based\cite{skreta2025feynmankac} and RNE-based~\cite{he2026rne} methods fail to recover the target mixture imbalance, instead remaining close to the bias of the proposal distribution. In contrast, FP++ is able to accurately recover the correct mode proportion.

\paragraph{Discussion.}
This behavior can be explained by the fact that FKC and RNE methods do not explicitly compute absolute density ratios. As a result, their induced distributions remain closely tied to the proposal distribution, rather than correcting for the true target imbalance.

In contrast, FP++ incorporates the Jacobian determinant term, which explicitly accounts for volume changes under the transformation. This enables FP++ to correctly reweight the induced distribution and recover the target mixture proportions independent of the proposal bias.

\subsection{1000D analytical Jacobian benchmark}

We further validate FP++ on a fully analytical 1000D Gaussian Mixture Model, where both the velocity field and its Jacobian determinant are available in closed form. This setting provides an exact reference and removes learned-model error from the Jacobian-estimation comparison.

\paragraph{Setup.}
We randomly sample initial points and evaluate the log-Jacobian determinant along the continuous trajectory $t \in [0,1]$. At each time step, results are averaged over multiple samples to reduce statistical noise.

\begin{figure}[htbp]
\centering
\includegraphics[width=0.75\textwidth]{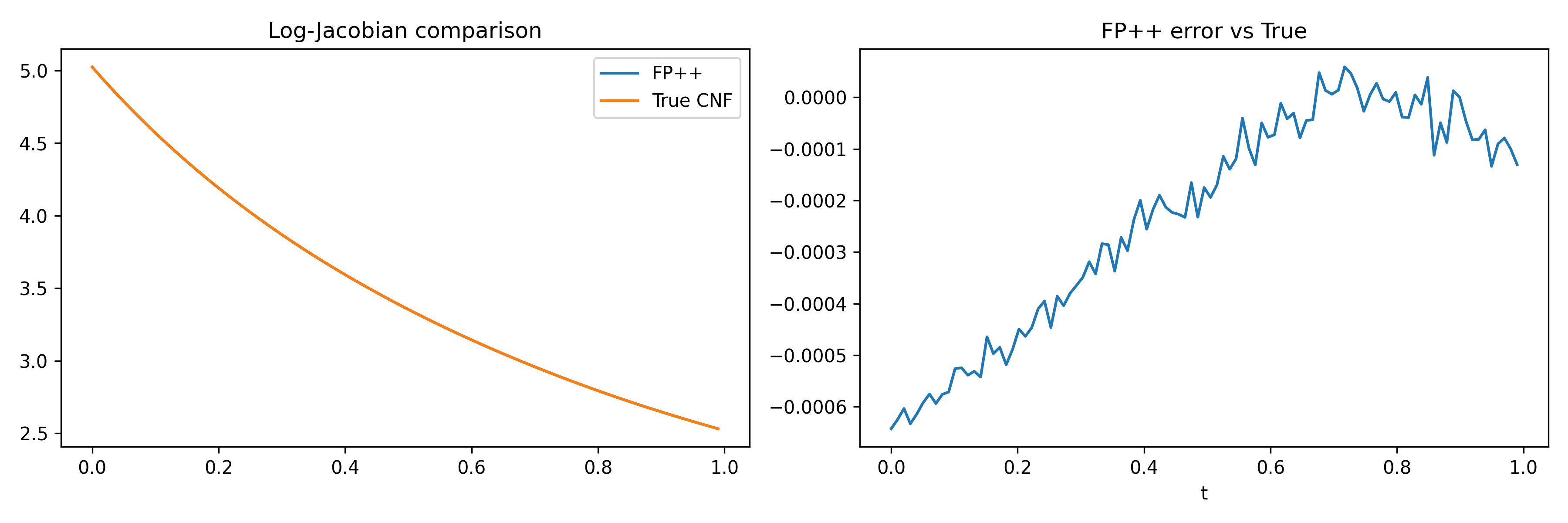}

\vspace{0.5em}

\caption{Jacobian determinant accuracy under a fully analytical 1000D Gaussian Mixture Model. FP++ closely matches the ground-truth log-Jacobian determinant over the entire trajectory $t \in [0,1]$.}
\label{fig:fp_analytic}
\end{figure}

\paragraph{Results.}
Figure~\ref{fig:fp_analytic} shows the evolution of the log-Jacobian determinant over time. We observe that FP++ closely matches the analytical ground truth throughout the entire trajectory, maintaining high accuracy even in 1000-dimensional space.

\paragraph{Conclusion.}
This experiment confirms that FP++ provides numerically consistent and accurate Jacobian determinant estimation under fully analytical conditions.

\subsection{FP++ Jacobian estimation with TBG pretrained models}
\label{app:tbg_ad2}

We further evaluate FP++ on a realistic molecular system using a pretrained Transferable Boltzmann Generators (TBG) model~\cite{klein2024transferable}. 
We consider the alanine dipeptide system as a benchmark for thermodynamic consistency and free energy estimation.

TBG is trained via equivariant continuous normalizing flows and can generate molecular configurations across different temperatures. However, its induced distribution generally deviates from the true Boltzmann distribution, leading to biased thermodynamic estimates.

FP++ addresses this issue by providing a low-cost determinant estimator along the flow trajectory. Its ideal exact-inverse construction is unbiased at the determinant level, while finite-difference and numerical-inversion effects make the practical implementation approximate. This enables importance reweighting of generated samples without retraining the generative model.

\begin{figure}[htbp]
\centering
\includegraphics[width=0.95\textwidth]{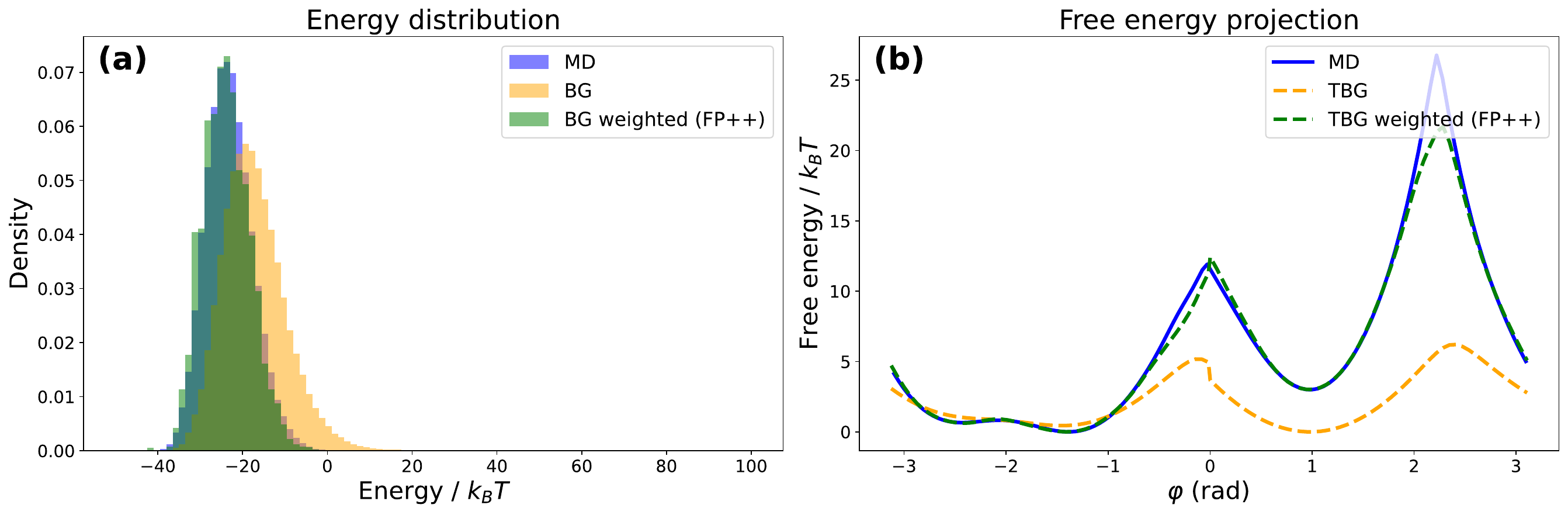}

\vspace{0.5em}

\caption{
Energy distribution and free energy profiles for the alanine dipeptide system. 
(a) Energy histograms comparing MD reference, unweighted TBG samples, and FP++-weighted samples. 
(b) Free energy profiles along the dihedral angle $\phi$, computed from MD, unweighted TBG, and FP++-weighted samples. The sample size is 200,000 configurations.
}
\label{fig:ala_fp++_results}
\end{figure}

We draw 200,000 independent configurations from the pretrained TBG model (denoted as unweighted samples). For each sample, FP++ estimates the multiplicative determinant correction, which is accumulated in log space to form importance weights $\exp(-W)$. In our implementation, the underlying ODE is solved using a 50-step fourth-order Runge--Kutta (RK4) integrator, and we use the FP3++ estimator with three probes per step. These weights are then used to construct a reweighted distribution (``FP++ weighted'').

We evaluate two key thermodynamic statistics:

\begin{enumerate}
    \item \textbf{Energy distribution.}
    Figure~\ref{fig:ala_fp++_results}(a) shows energy histograms for MD simulations, unweighted TBG samples, and FP++-weighted samples. The unweighted TBG distribution exhibits a clear mismatch with the MD reference, whereas FP++ weighting significantly corrects this discrepancy, closely matching the ground-truth energy distribution.

    \item \textbf{Free energy profile.}
    We compute the free energy landscape along the dihedral angle $\phi$, the slowest collective variable of alanine dipeptide. Kernel density estimation is used to estimate $P(\phi)$, followed by $F(\phi) = -k_B T \ln P(\phi)$. As shown in Figure~\ref{fig:ala_fp++_results}(b), unweighted samples exhibit biased free energy barriers and incorrect relative basin depths, while FP++-weighted results closely align with MD simulations, including accurate recovery of transition regions.
\end{enumerate}

Overall, these results demonstrate that FP++ enables accurate thermodynamic reweighting of pretrained generative models without retraining, providing a principled correction mechanism for Jacobian-induced distributional bias in molecular simulation.

\begin{table}[ht]
\centering
\caption{Alanine-dipeptide direct-reweighting diagnostics. Arrows report the metric before and after weighting.}
\label{tab:ad2_quality}
\small
\setlength{\tabcolsep}{6pt}
\begin{tabular}{lcc}
\toprule
Metric & FP++ direct & TBG official \\
\midrule
Valid aligned samples & 199885 & 199871 \\
ESS & 3309.496 & 12854.656 \\
Normalized ESS & 0.016557 & 0.064315 \\
Energy-W2 (unw.$\rightarrow$w.) & $6.175459\rightarrow0.709800$ & $6.714405\rightarrow0.794129$ \\
$\phi$ torsion-W2 (unw.$\rightarrow$w.) & $1.438921\rightarrow0.009786$ & $1.446999\rightarrow0.006618$ \\
$\psi$ torsion-W2 (unw.$\rightarrow$w.) & $0.752541\rightarrow0.046143$ & $0.760560\rightarrow0.027939$ \\
Normalized weight variance & $1.486644\times10^{-9}$ & $3.641829\times10^{-10}$ \\
Normalized weight CV$^2$ & 59.397412 & 14.548529 \\
Log-weight variance & 12.548962 & 11.570547 \\
\bottomrule
\end{tabular}
\end{table}

\begin{table}[ht]
\centering
\caption{Jacobian-computation cost on 100 alanine-dipeptide samples.}
\label{tab:ad2_cost}
\small
\begin{tabular}{lcccc}
\toprule
Method & Time (s) & s/sample & Alloc. (MB) & Reserved (MB) \\
\midrule
FP3++ & 80.511 & 0.805 & 608.22 & 676.00 \\
Exact Jacobian & 569.240 & 5.692 & 4276.49 & 6614.00 \\
\bottomrule
\end{tabular}
\end{table}

Tables~\ref{tab:ad2_quality} and~\ref{tab:ad2_cost} summarize the trade-off between computational cost and reweighting quality. Official TBG weighting has higher normalized ESS and lower weight CV$^2$, whereas FP++ gives slightly lower Energy-W2 and torsion-W2 of comparable magnitude while reducing exact-Jacobian cost: FP3++ is $7.07\times$ faster, with $7.03\times$ lower peak allocated memory and $9.78\times$ lower peak reserved memory. The idealized pass-count ratio is $67/7=9.57\times$ because three central-difference probes require six perturbed evaluations in addition to the forward solve; solver and memory overhead account for the smaller observed speedup.

\subsection{Reweighting the Ace-AAA-Nme system}
\label{app:al4_ecnf}
We evaluate reweighting on a 42-atom Ace-AAA-Nme system (126 coordinate dimensions) using an ECNF++ generator. The same 2,000 generated samples are reweighted with either the exact Jacobian or FP5++. Results are shown in Table~\ref{tab:al4_ecnf}; exact-Jacobian weighting yields higher ESS and lower Energy-W2, whereas FP5++ yields lower torsion-joint W2.
\begin{table}[ht]
\centering
\caption{AL4\_ECNF (Ace-AAA-Nme, 300 K) direct-reweighting metrics.}
\label{tab:al4_ecnf}
\small
\begin{tabular}{lccc}
\toprule
Reweighting & ESS/N & Energy-W2 & Torsion-joint W2 \\
\midrule
Exact Jacobian & 0.006896 & 6.0864 & 3.1380 \\
FP5++ & 0.000620 & 9.4396 & 2.7036 \\
\bottomrule
\end{tabular}
\end{table}

\subsection{Key hyperparameters for reproducibility}
\label{app:hyperparams}

Several hyperparameters are critical for reproducing the variance reduction observed in FP and FP++ methods.  
The exact hyperparameter values and full implementation details are available in the public code repository listed under Software and Data in the main text.

%%%%%%%%%%%%%%%%%%%%%%%%%%%%%%%%%%%%%%%%%%%%%%%%%%%%%%%%%%%%

\section{Impact statement}
\label{app:impact}
FP++ provides a general-purpose, determinant-unbiased alternative (for exact inverse-step maps and directional derivatives) to Hutchinson trace estimators for computing Jacobian determinants in continuous generative flows.  
By reducing variance under the conditions in Appendix~\ref{sec:fpplusplus_variance} and supporting multi-noise extensions, it enables rigorous thermodynamic inference and Boltzmann reweighting from pretrained generative models.  
This work advances the use of flow-based generative models in scientific applications, including molecular simulations and high-dimensional inference, without raising immediate ethical concerns.

\section{Statement on the use of large language models}
\label{app:llm_statement}

During the preparation of this manuscript, large language models (LLMs) were used in a limited manner
solely for language editing purposes, such as improving clarity, grammar, and academic style.
All aspects of the research conception, methodological development, experimental design,
analysis of results, and the scientific conclusions presented in this paper
were carried out independently by the authors.

%%%%%%%%%%%%%%%%%%%%%%%%%%%%%%%%%%%%%%%%%%%%%%%%%%%%%%%%%%%%

\newpage
\section*{NeurIPS Paper Checklist}

\begin{enumerate}

\item {\bf Claims}
    \item[] Question: Do the main claims made in the abstract and introduction accurately reflect the paper's contributions and scope?
    \item[] Answer: \answerYes{} % Replace by \answerYes{}, \answerNo{}, or \answerNA{}.
    \item[] Justification: The abstract summarizes the ideal estimator and tested applications. Section \ref{sec:method} and Appendices \ref{sec:fp_pp}--\ref{sec:fpplusplus_variance} specify the exact-inverse assumptions, the numerical approximation, and the conditions for variance reduction; the empirical scope is stated in the results and limitations.

\item {\bf Limitations}
    \item[] Question: Does the paper discuss the limitations of the work performed by the authors?
    \item[] Answer: \answerYes{} % Replace by \answerYes{}, \answerNo{}, or \answerNA{}.
    \item[] Justification: The limitations paragraph and Appendix \ref{app:fd} discuss finite-difference, roundoff, solver-inversion, covariance-condition, and scope limitations.

\item {\bf Theory assumptions and proofs}
    \item[] Question: For each theoretical result, does the paper provide the full set of assumptions and a complete (and correct) proof?
    \item[] Answer: \answerYes{} % Replace by \answerYes{}, \answerNo{}, or \answerNA{}.
    \item[] Justification: Appendix \ref{sec:fp_pp} states the exact-inverse assumptions for determinant unbiasedness. Appendix \ref{sec:fpplusplus_variance} gives the covariance decomposition and sufficient conditions rather than claiming universal variance reduction.

    \item {\bf Experimental result reproducibility}
    \item[] Question: Does the paper fully disclose all the information needed to reproduce the main experimental results of the paper to the extent that it affects the main claims and/or conclusions of the paper (regardless of whether the code and data are provided or not)?
    \item[] Answer: \answerYes{} % Replace by \answerYes{}, \answerNo{}, or \answerNA{}.
    \item[] Justification: Section \ref{sec:method}, Appendix \ref{app:smc_protocol}, and Appendix \ref{app:hyperparams} specify the determinant estimator, particle counts, annealing schedules, resampling rule, mutation steps, and key numerical settings.

\item {\bf Open access to data and code}
    \item[] Question: Does the paper provide open access to the data and code, with sufficient instructions to faithfully reproduce the main experimental results, as described in supplemental material?
    \item[] Answer: \answerYes{} % Replace by \answerYes{}, \answerNo{}, or \answerNA{}.
    \item[] Justification: The public repository linked in Section \ref{sec:software_data} provides the implementation, environment requirements, and training, MC, and SMC usage examples; externally released molecular data and pretrained models are cited in the paper.

\item {\bf Experimental setting/details}
    \item[] Question: Does the paper specify all the training and test details (e.g., data splits, hyperparameters, how they were chosen, type of optimizer) necessary to understand the results?
    \item[] Answer: \answerYes{} % Replace by \answerYes{}, \answerNo{}, or \answerNA{}.
    \item[] Justification: Training and evaluation details for the 1000D GMM and Chignolin mutant are specified in Section \ref{sec:results} and Appendix \ref{app:hyperparams}, including annealing schedules, MCMC steps, and force field parameters.

\item {\bf Experiment statistical significance}
    \item[] Question: Does the paper report error bars suitably and correctly defined or other appropriate information about the statistical significance of the experiments?
    \item[] Answer: \answerYes{} % Replace by \answerYes{}, \answerNo{}, or \answerNA{}.
    \item[] Justification: Tables \ref{table:GMM10d_results}, \ref{tab:compare_GMM}, and \ref{table:compare_GCN} report means and standard deviations over ten independent runs for FP/FP++ and Hutchinson estimators. The reduced-particle, single-run 1000D BFJacob reference is explicitly identified. Standard deviations quantify run-to-run variability rather than constituting a formal significance test; representative-run diagnostics are labeled separately.

\item {\bf Experiments compute resources}
    \item[] Question: For each experiment, does the paper provide sufficient information on the computer resources (type of compute workers, memory, time of execution) needed to reproduce the experiments?
    \item[] Answer: \answerNo{} % Replace by \answerYes{}, \answerNo{}, or \answerNA{}.
    \item[] Justification: We report ODE-equivalent passes, per-run wall-clock times, and the alanine-dipeptide runtime and peak-memory benchmark, but the complete hardware inventory and total project-wide compute (including exploratory runs) are not fully reported for every experiment.
    
\item {\bf Code of ethics}
    \item[] Question: Does the research conducted in the paper conform, in every respect, with the NeurIPS Code of Ethics \url{https://neurips.cc/public/EthicsGuidelines}?
    \item[] Answer: \answerYes{} % Replace by \answerYes{}, \answerNo{}, or \answerNA{}.
    \item[] Justification: Our work adheres to the NeurIPS Code of Ethics regarding scientific integrity and responsible data usage in molecular modeling research.

\item {\bf Broader impacts}
    \item[] Question: Does the paper discuss both potential positive societal impacts and negative societal impacts of the work performed?
    \item[] Answer: \answerYes{} % Replace by \answerYes{}, \answerNo{}, or \answerNA{}.
    \item[] Justification: The impact statement in Appendix \ref{app:impact} discusses scientific benefits for molecular sampling and high-dimensional inference; the authors do not identify immediate ethical concerns for these applications.
    
\item {\bf Safeguards}
    \item[] Question: Does the paper describe safeguards that have been put in place for responsible release of data or models that have a high risk for misuse (e.g., pre-trained language models, image generators, or scraped datasets)?
    \item[] Answer: \answerNA{} % Replace by \answerYes{}, \answerNo{}, or \answerNA{}.
    \item[] Justification: The proposed method and datasets (GMM and proteins) are scientific in nature and do not pose a high risk of malicious misuse or dual-use.

\item {\bf Licenses for existing assets}
    \item[] Question: Are the creators or original owners of assets (e.g., code, data, models), used in the paper, properly credited and are the license and terms of use explicitly mentioned and properly respected?
    \item[] Answer: \answerNo{} % Replace by \answerYes{}, \answerNo{}, or \answerNA{}.
    \item[] Justification: The sources of the force-field parameters, molecular data, pretrained flows, and comparison models are credited. However, a complete asset-by-asset inventory of licenses and terms of use is not explicitly provided; we therefore do not claim that all assets have an academic or MIT license.

\item {\bf New assets}
    \item[] Question: Are new assets introduced in the paper well documented and is the documentation provided alongside the assets?
    \item[] Answer: \answerYes{} % Replace by \answerYes{}, \answerNo{}, or \answerNA{}.
    \item[] Justification: The FP++ implementation is released in our repository alongside environment requirements and training, MC, and SMC usage examples.

\item {\bf Crowdsourcing and research with human subjects}
    \item[] Question: For crowdsourcing experiments and research with human subjects, does the paper include the full text of instructions given to participants and screenshots, if applicable, as well as details about compensation (if any)? 
    \item[] Answer: \answerNA{} % Replace by \answerYes{}, \answerNo{}, or \answerNA{}.
    \item[] Justification: This research does not involve crowdsourcing or human subjects.

\item {\bf Institutional review board (IRB) approvals or equivalent for research with human subjects}
    \item[] Question: Does the paper describe potential risks incurred by study participants, whether such risks were disclosed to the subjects, and whether Institutional Review Board (IRB) approvals (or an equivalent approval/review based on the requirements of your country or institution) were obtained?
    \item[] Answer: \answerNA{} % Replace by \answerYes{}, \answerNo{}, or \answerNA{}.
    \item[] Justification: Our study is limited to physical simulations and mathematical modeling; no human subject research was performed.

\item {\bf Declaration of LLM usage}
    \item[] Question: Does the paper describe the usage of LLMs if it is an important, original, or non-standard component of the core methods in this research? Note that if the LLM is used only for writing, editing, or formatting purposes and does \emph{not} impact the core methodology, scientific rigor, or originality of the research, declaration is not required.
    %this research? 
    \item[] Answer: \answerNA{} % Replace by \answerYes{}, \answerNo{}, or \answerNA{}.
    \item[] Justification: LLMs were not an important, original, or non-standard component of the core research method; any use was limited to writing, editing, or formatting, for which the NeurIPS question states that a declaration is not required.

\end{enumerate}

\end{document}